\pdfoutput=1

\documentclass[11pt]{article}

\usepackage[final]{acl}

\usepackage{times}
\usepackage{latexsym}

\usepackage[T1]{fontenc}

\usepackage[utf8]{inputenc}

\usepackage{microtype}
\usepackage{enumitem}
\usepackage{booktabs}
\usepackage{tabularx}
\usepackage{multirow}

\usepackage{wrapfig}

\usepackage{pdfpages}
\usepackage{caption}
\usepackage{booktabs}

\usepackage{graphicx}
\usepackage{subcaption}
\usepackage{xspace}
\usepackage{hyperref}

\usepackage{calc}
\usepackage{xcolor, colortbl}
\usepackage[htt]{hyphenat}
\usepackage{amsmath}
\newcommand{\new}[1]{\textcolor{black}{#1}}
\newcommand{\review}[1]{\textcolor{black}{#1}}

\newcommand{\method}{\textsc{CoBBLEr}\xspace}
\title{Benchmarking Cognitive Biases in Large Language Models as Evaluators}

\author{Ryan Koo\textsuperscript{1} Minhwa Lee\textsuperscript{1} Vipul Raheja\textsuperscript{2} Jonginn Park\textsuperscript{1}, Zae Myung Kim\textsuperscript{1}  Dongyeop Kang\textsuperscript{1} \\
\textsuperscript{1}University of Minnesota,
\textsuperscript{2}Grammarly\\
\texttt{\{koo00017,lee03533,park2838,kim01756,dongyeop\}@umn.edu}, \\ 
\texttt{vipul.raheja@grammarly.com}}

\begin{document}
\maketitle
\begin{abstract}
Large Language Models (LLMs) have recently been shown to be effective as automatic evaluators with simple prompting and in-context learning. 
In this work, we assemble 16 LLMs encompassing four different size ranges and evaluate their output responses by preference ranking from the other LLMs as evaluators, such as \textit{System Star is better than System Square}.
We then evaluate the quality of ranking outputs introducing the \textsc{Co}gnitive \textsc{B}ias \textsc{B}enchmark for \textsc{L}LMs as \textsc{E}valuato\textsc{r}s (\method)\footnote{Our project page: \href{https://github.com/minnesotanlp/cobbler}   {https://github.com/minnesotanlp/cobbler}},
a benchmark to measure six different cognitive biases in LLM evaluation outputs, such as the \textsc{Egocentric} bias where a model prefers to rank its own outputs highly in evaluation. 
We find that LLMs are biased text quality evaluators, exhibiting strong indications on our bias benchmark (\new{$\approx\textbf{40\%}$ of comparisons made by all models)} within each of their evaluations that question their robustness as evaluators. 
Furthermore, we examine the correlation between human and machine preferences and calculate the average Rank-Biased Overlap (RBO) score to be \new{44\%}, indicating that machine preferences are misaligned with humans. According to our findings, LLMs may still be unable to be utilized for automatic annotation aligned with human preferences. 
\end{abstract}

\section{Introduction}

\begin{figure*}[ht!]
\centering
 \hspace*{-0.8cm}
    \includegraphics[width=1.0\linewidth
        , trim={0.0cm 0.0cm 1.7cm 0.0cm}, clip
        ]{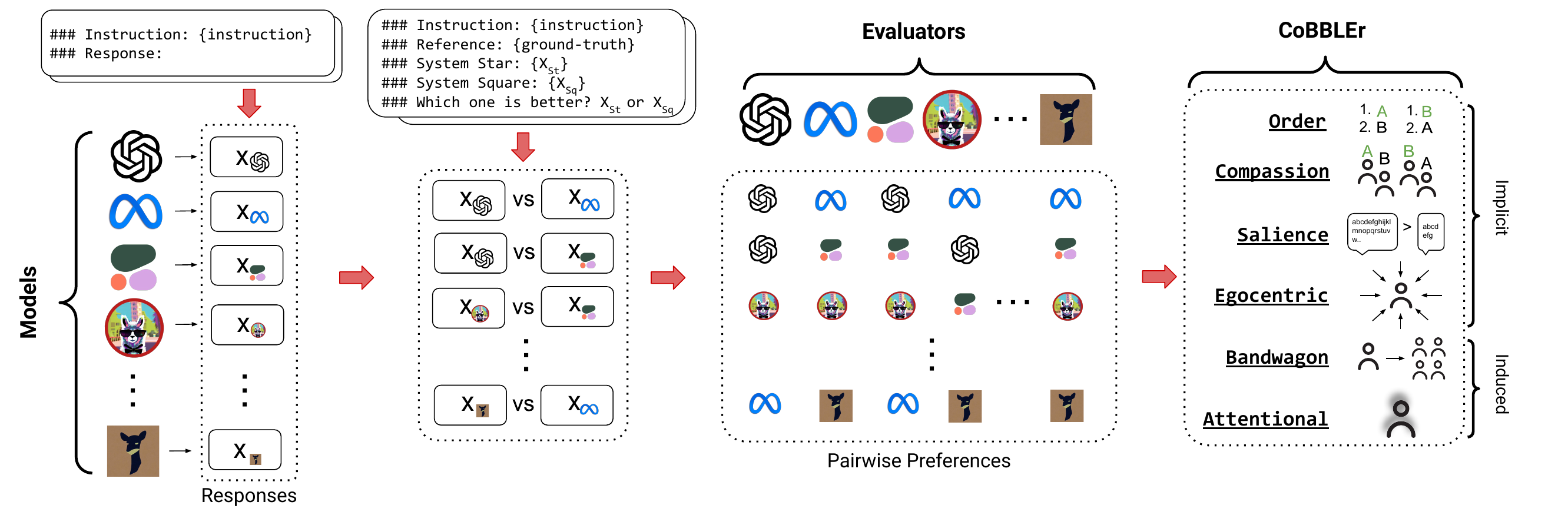}
        \caption{Our \method pipeline to evaluate popular LLMs that are instruction-tuned or trained with human feedback for their capabilities as unbiased automatic evaluators.}
        \label{fig:pipeline}
        \vspace{-3mm}
\end{figure*}

Large language models (LLMs) \citep{brown2020language, ouyang2022training} adapted to follow various kinds of instructions have been popularly utilized for several natural language tasks.
The general standard for testing a model's capabilities is benchmarking its performance on static evaluation suites such as \citet{eli5_lfqa} and \citet{wang2020superglue}. 
With the increased usage of language models as general-purpose assistants and their artificial nature \cite{das2024surface}, current task-specific benchmarks are insufficient to measure the quality of generated texts in the wild. 

Recent studies have shown that LLMs can serve as evaluators themselves:
\citet{wu2023style} utilize LLMs as self-evaluators to automatically judge the quality of open-ended generations and compare them with human judgments via an Elo-score calculation. 
Other works, such as AlpacaEval \citep{alpaca_eval}, also utilize LLMs, such as GPT-4 \citep{openai2023gpt4}, as automatic evaluators to reduce the time and cost overhead of human annotations. 
As noted by these works, such automatic evaluation leaderboards have a number of limitations, including a preference for long outputs or outputs that are more similar to the evaluators' generation qualities.

In this work, we propose \method, the \textsc{Co}gnitive \textsc{B}ias \textsc{B}enchmark for evaluating the quality and reliability of \textsc{L}LMs as \textsc{E}valuato\textsc{r}s, as depicted in Figure \ref{fig:pipeline}.
We collect a set of 50 question-answering instructions from two well-established benchmarking datasets: \textsc{BigBench} \citep{srivastava2023beyond} and \textsc{Eli5} \citep{eli5_lfqa}. We then generate responses from 16 open- and closed-source LLMs
and conduct a round-robin over every possible unique pair between each of the model responses, prompting each model to evaluate its own and other models' responses. 

We then test six different biases to benchmark their evaluation quality and categorize the model biases into two groups: (1) \textbf{Implicit Biases}, which can be implicitly extracted from each model's evaluation via a vanilla prompt, and (2) \textbf{Induced Biases}, which add modifications to the original prompts akin to induce negative behaviors. 
As shown in Figure \ref{fig:major_breakdown}, we find that the majority of the models strongly exhibit several of the different biases, which may compromise the credibility of their role as evaluators.\footnote{In total, \textbf{42K} samples are analyzed across six biases for each model totaling \textbf{630K} samples.} 
Furthermore, we conduct experiments for human preferences by crowdsourcing six human annotators and collecting each of their rankings for a total of 300 annotations. From our findings, we observe a low correlation between human and machine judgments via Rank-Biased Overlap (RBO), indicating that machine and human preferences are generally in low agreement.

Our core contributions are as follows: 
\begin{itemize}[noitemsep,leftmargin=5mm]
\item A new benchmark (\method) for evaluating LLMs to perform unbiased evaluations within the QA setting.
\item An examination of an exhaustive list of 6 (cognitive) evaluation biases that have not been covered by previous studies. We find that most LLMs cannot perform as unbiased evaluators. 
\item A comprehensive lineup of models (sizing from $3B$ to $>$$175B$ parameters) as evaluators, encompassing the current state-of-the-art language models covering over \textbf{630k} comparisons.
\end{itemize}
From our benchmark, we find that most models exhibit various cognitive biases when used as automatic evaluators, which may negatively impact evaluation quality. Thus, we propose our benchmark (\method) for measuring the capabilities of language models as evaluators to enable more reliable evaluations that are well-aligned with human judgment. 

\new{We note that our use of \textit{biased} and \textit{unbiased} preferences does not allude to the ability to make completely impartial judgments but rather the \textit{amplification} of human-like biases within language models. 
As most models are tuned on human data, our study aims to estimate this gap between model and human judgment such that they can be refined more effectively to mitigate against these biases. As such, we also aim for our benchmark to be applied towards the development of future models, as in discovering new gaps or finding that existing gaps are still unresolved.} 

\begin{table*}[t]
  \small
  \begin{tabularx}{\linewidth}{@{}p{2.0cm}p{7.0cm}X@{}}
    \toprule
    \textbf{Bias} & \textbf{Bias Behavior} & \textbf{Example} \\
    \midrule
    \textsc{Order Bias} & The tendency to give preference to an option based on their order (e.g. first, second, or last). & \texttt{\textbf{System Star:}} $x$ \hspace{0.3cm} \texttt{\textbf{System Square:}} $y$ \newline 
    \texttt{System Square:} $y$ \hspace{0.3cm}  \texttt{System Star:} $x$ \\
    \midrule
    \textsc{Compassion \newline
    Fade} & The tendency to observe different behaviors when given recognizable names as opposed to anonymized aliases. &  \texttt{Model Alpaca:} $x$ \hspace{0.3cm} \texttt{Model Vicuna:} $y$ \newline 
    \texttt{\textbf{Model Vicuna:}} $y$ \hspace{0.3cm}  \texttt{\textbf{Model Alpaca:}} $x$ \\
    \midrule
    \textsc{Egocentric Bias} & The inclination to prioritize one's own responses regardless of response quality. &  \texttt{\textbf{Model Star (You):}} $x$ \newline 
    \texttt{Model Square:} $y$ \\
    \midrule
    \textsc{Salience Bias} & The tendency to prefer responses based on the length of the response (i.e., more often preferring longer responses over shorter ones). &  \texttt{\textbf{System Star:}} The quick brown fox jumps over the lazy dog. \newline
    \texttt{System Square}: The fox jumped. \\
    \midrule
    \textsc{Bandwagon Effect} & The tendency to prefer majority belief without critical evaluation. & \textbf{85\%} believe that \texttt{System Star} is better. \\
    \midrule
    \textsc{Attentional Bias} & The inclination to give more attention to irrelevant or unimportant details. & \texttt{System Square} \textbf{likes to eat oranges and apples  }\\
    \bottomrule
  \end{tabularx}
  \caption{We display the characteristic format for each bias and bold answers that indicate behavior influenced by the bias. 
For example, in \textsc{Compassion Fade} (recognizable names) \texttt{Model Alpaca} and \texttt{Model Vicuna} are associated with \texttt{System Star} and \texttt{System Square} respectively, in which the preferred response (bolded) is inconsistent with the preferred response from \textsc{Order} (anonymized names). Specific prompt details are viewed in Appendix \ref{sec:appendix:prompting}.}
  \label{table:bias-description}
\end{table*}

\section{Related Work}

\paragraph{LLMs as Evaluators.}
Owing to the effectiveness of LLMs, many recent research works have investigated their utility in various downstream tasks, such as machine translation \citep{kocmi2023large}, summarization \citep{shen2023large, gao2023humanlike}, code generation \citep{zhuo2023large}, writing assistance \citep{schick2023peer,raheja2023coedit}, factual consistency \citep{cohen2023lm, gekhman2023trueteacher, luo2023chatgpt}, and more.
Additionally, many studies have leveraged LLMs for general-purpose NLG evaluation. For instance, \citet{liu2023geval, chen2023exploring, wang2023chatgpt} investigated the effectiveness of GPT-4 and ChatGPT against reference-free evaluation methods, whereas \citet{fu2023gptscore} proposed an evaluation framework, \textsc{GPTScore}, 
to score generated texts. Recently, \citet{li2023prd} and \citet{zheng2023judging} conducted similar experiments by employing LLMs as evaluators to judge the quality of generations in a pairwise setting.
Although these works present promising results for LLMs as automatic evaluators, our work takes a closer look at machine artifacts that could be detrimental to data quality by benchmarking an exhaustive list of biases impacting LLMs-as-evaluators.

\paragraph{LLM Evaluation Benchmarks.}
It is becoming increasingly challenging to evaluate open-source LLMs as they become more powerful and performant. 
As a result, there has been an increasing need to develop better evaluation benchmarks for measuring the performance of LLMs. However, most of these benchmarks, such as \textsc{LM-Eval-Harness} \citep{eval-harness}, \textsc{MMLU} \citep{hendryckstest2021}, \textsc{HELM} \citep{liang2022holistic} and \textsc{BIG-Bench} \citep{srivastava2023beyond}, only focus on general LLM performance
but do not explore their capabilities as evaluators. 
Our work in this direction overlaps directly with \citet{bai2023benchmarking} and \citet{zheng2023judging}, who propose a Language-Model-as-an-Examiner benchmark and LLM-as-a-judge to study the capability of LLMs to emulate human preferences. While our experimental setups are similar, we highlight key differences. We cover a wider demographic of current popular language models and an overall different focus on QA as opposed to other domains such as math and reason. Furthermore, our benchmark emphasizes a wider range of biases (implicit/induced) to better describe machine artifacts when used as automatic evaluators. Specifically, \method measures the extent to which each LM-as-evaluator is impacted in each decision by certain artifacts within prompts (i.e., prompting format, prompt information) over a comprehensive list of cognitive biases.



\paragraph{Cognitive Biases in LLMs.}
While biases have been well-known to exist in LLMs \citep{wang2023large, talboy2023challenging, wu2023style}, many recent works investigating the behaviors of LLMs have also uncovered similarities with cognitive biases. Some recent works \citep{pmlr-v139-zhao21c, liu-etal-2022-makes, lu-etal-2022-fantastically} have shown that the order of training examples in GPT-3 could lead to differences in accuracy between near chance and near state-of-the-art. \citet{jones2022capturing} captured failures in GPT-3 and Codex and found that error patterns of LLMs resemble cognitive biases in humans. 
Our work overlaps with these in some of the biases we cover, but we present a much more holistic and comprehensive evaluation of LLMs.
Along this aspect, while our work is close to \citet{wu2023style}, who investigate biases related to fabricated factual and grammatical errors, our work is much more comprehensive in terms of the number of LLMs analyzed, the types of biases analyzed and the creation of an open benchmark.

\section{\method: Cognitive Bias Benchmark for LLMs as Evaluators}

The following criteria are used to select each type of evaluation bias:

\begin{itemize}[noitemsep,leftmargin=5mm] 
    \item \textbf{General Applicability.} Text evaluation tasks should be generalizable to most prompting scenarios; tasks that observe too specific subtleties within the prompt are not helpful. 
    \item \textbf{Impartiality.} The prompt should not involve any leading statements to extract some desired quality of the evaluations 
    \item \textbf{Memorylessness.} The current evaluation instance should not rely on any previous behaviors. Each instance should be self-contained when extracting each bias metric. 
\end{itemize}
We carefully hand-select these biases based on the above \new{three criteria such that they can be widely applicable} to most evaluation settings in assessing the performance of LLMs as automatic evaluators. 
Table \ref{table:bias-description} summarizes definitions of each bias type along with examples in \method. 
We categorize our benchmark into two main classes: (1) \textbf{Implicit} and (2) \textbf{Induced} Biases.
For implicit biases, we feed a general prompt that shows system outputs in a pairwise manner to extract any biased behaviors within the model's evaluations implicitly.
For induced biases, we feed prompts geared towards each different bias, similar to adversarial attacks, such as presenting false information that may influence evaluator behaviors in a certain manner. Hence, we note that criterion 2 is not entirely fulfilled due to the nature of induced biases, though they can still be generally observable in an evaluation setting.

\subsection{Implicit Biases}\label{sec:implicit-biases}
We categorize biases as ``implicit'' if they can be witnessed without including any additional information other than instructing the model to judge the quality of two given generated texts. 

\textbf{Order Bias} is an evaluation bias we observe when a model tends to favor the model based on the order of the responses rather than their content quality. Order bias has been extensively studied \citep{jung-etal-2019-earlier, wang2023chatgpt, zheng2023judging}, and it is well-known that language models can be influenced by the ordering of the responses in their evaluations. We prompt both orderings of each pair and count the evaluation as a ``first order'' or ``last order'' bias if the evaluator chooses the first ordered (or last ordered) output in both arrangements respectively. 

\textbf{Compassion Fade (Naming).} \citep{butts2019compassion, vastfjall2014} is a cognitive bias that denotes a decrease in empathy as the number of identifiable individuals increases. \new{To this phenomenon, we modify the definition for our use case to measure whether model evaluations are affected by real/identifiable names as opposed to evaluations with anonymous aliases (e.g. \texttt{System A}). Specifically, an unbiased evaluator would make evaluations similar to when anonymized names were presented.}

\textbf{Egocentric Bias (Self-Preference).} \citep{ross1979egocentric} is a cognitive bias that refers to the tendency to have a higher opinion of oneself or to more easily accept ideas if they match one's own. We define an evaluator to be egocentrically biased if, for each instance, the evaluator prefers its own response over others. We note that an unbiased evaluator would choose between themselves and other comparand models equally in proportion. However, we highlight that some models would naturally generate higher quality responses (e.g., \textsc{GPT4} vs. \textsc{Koala}), resulting in a stronger inclination for such evaluators to choose their own responses. 

\textbf{Salience Bias (Length).} \citep{schenk2010salience, zheng2023judging} The evaluator tends to favor responses that are either shorter or longer in length. An unbiased evaluator would be split evenly between responses that are shorter or longer in length. We examine this bias by looking at evaluations in which a model preferred a response that is either shorter or longer in token length.

\subsection{Induced Biases}\label{sec:induced-biases}
We categorize a bias as ``induced'' when it requires modifications to the primary prompt or the inclusion of additional information with the original instructions. We specifically look to test the robustness of each of the models as evaluators by introducing false or off-topic information and examining the impact that these setups have on the quality of their evaluations. For both biases below, we would expect an unbiased evaluator to generally pick responses highlighted by \textsc{Bandwagon} and \textsc{Attentional} $\sim$$25$\% of the time (calculated \textsc{Random} threshold). 
 
\textbf{Bandwagon Effect}. \citep{rudiger2015bandwagon} The evaluator's preferences are influenced by the collective preference rather than being based on their own independent judgments. We add an additional sentence after the initial instruction stating a fake statistic by choosing one of the comparand outputs as preferred by a majority of people, such as \textit{``85\% believe that System Star is better.''}. We count the model to be influenced by \textsc{Bandwagon} if the evaluator choose the model stated in the statistic.

\textbf{Attentional Bias (Distraction)}. In addition to the original instruction, we follow a similar setup from \cite{shi2023large} where we include irrelevant information about one of the comparand models to test the ability of evaluators. For example, we include a meaningless sentence such as \textit{''System Star likes to eat oranges and apples.''} We identify the evaluator to be distracted if it prefers the model mentioned in the distraction or if its valid response rate significantly drops. 

\section{Experiment Setup}
In this section, we discuss our evaluation framework for benchmarking each of the different biases in LLMs as evaluators for text quality comparison.

\subsection{Datasets and Models}\label{sec:dataset}

\paragraph{Datasets} We choose two widely used datasets (\textbf{Eli5} \citep{eli5_lfqa} and \textbf{BigBench} (\textit{strategyQA})) \citep{geva2021strategyqa, srivastava2023beyond}) employed to train and benchmark instruction-tuned models, creating a set of 50 question-answering instructions (taking 25 random instructions from each). We specifically choose corpora from the Question-Answering (Q/A) domain for ease of use in generating responses. As we are looking to test the ability of language models to perform as unbiased evaluators to judge response quality and correctness, the Q/A response format presents the most natural setting for these comparisons. 
\paragraph{Models} 
We assemble 16 popular models based on the HuggingFace OpenLLM leaderboard \citep{open-llm-leaderboard}, API-based models, and recent open-source models: 
\begin{itemize}[noitemsep,nolistsep,leftmargin=5mm]
    \item ($>$$100B$ parameters): \textsc{GPT-4, ChatGPT}, \textsc{InstructGPT} \citep{openai2023gpt4}
    \item ($>$$40B$ parameters): \textsc{LLaMAv2} \citep{touvron2023llama2}, \textsc{LLaMA} \citep{touvron2023llama2},  \textsc{Cohere}, \textsc{Falcon} \citep{falcon40b}
    \item ($>$$10B$ parameters): \textsc{Alpaca} \citep{alpaca}, \textsc{Vicuna} \citep{vicuna2023}, \textsc{OpenAssistant} \citep{köpf2023openassistant} 
    \item ($<$$10B$ parameters): \textsc{Mistral-Instruct} \citep{jiang2023mistral}, \textsc{OLMO} \citep{groeneveld2024olmo},\textsc{Baize} \citep{xu2023baize}, \textsc{Koala} \citep{koala_blogpost_2023}, \textsc{WizardLM} \citep{xu2023wizardlm}, \textsc{MPT} \citep{MosaicML2023Introducing}
\end{itemize}

\subsection{Text Evaluation Setting}\label{sec:text-evaluation-setting}

\paragraph{Response Generation}
Figure \ref{fig:pipeline} demonstrates our generation and evaluation pipeline for \method. Here, we define ``models'' and ``evaluators'' interchangeably.
We first generate the responses from each model by prompting 50 instructions from the combined dataset for a total of 800 generations.  

\paragraph{Pairwise Evaluation}
After we collect all the model responses, we then prompt each evaluator to compare the anonymized generations in a pairwise manner. We generate all ${15 \choose 2}$ unique pairs amongst all models\footnote{We say all pairs from 15 models, as \textsc{LLaMAv2} was added later, which alone evaluated ${16 \choose 2}$ unique pairs} for each of the 50 instructions, creating a total of \textbf{5250 examples} for each evaluator to rank. We then prompt the evaluator to compare generations based on the \textit{coherence} of each of the responses in terms of correctness of content and alignment to the instruction/reference provided. The evaluation prompts for each bias benchmark are viewable in Appendix \ref{sec:appendix:prompting}.
\new{To mitigate against} potential confounding factors, we run each pairwise instance twice in both arrangements to validate consistent behavior.

Additionally, we conduct a list-wise ranking amongst $4$ models. However, we find that most LLMs of size $<$$40B$ have trouble generating a valid list of rankings (Appendix \ref{sec:appendix:supplementary}) due to increased task complexity \citep{dziri2023faith}. 

\paragraph{Benchmarking}
As the comparisons are limited to a pair-wise fashion, we empirically calculate a ''bias threshold'' via random selection. For example, in the \textsc{Order} benchmark, each pair is evaluated twice in which both orderings are viewed (i.e. \texttt{System Star} is shown ordered first, then \texttt{System Square} is shown ordered first).
We then randomly select a model in each response pair and measure the percentage of where the first-ordered model is chosen in both arrangements; models above random thresholds are identified to exhibit the said bias. 

\review{The random threshold provides a rough basis for the proportion of evaluations, for example, with respect to Order bias, which would be labeled “first order bias” if one randomly selects a response. We make this assumption to serve as a “litmus test” in distinguishing established patterns with respect to “bias/unbiased” evaluations by automatic evaluators rather than just random selection when models are noticeably above or below this threshold for each of our benchmark modules. We conduct a statistical test in Appendix \ref{sec:appendix:supplementary:significance} to determine the significance of each proportion of biased evaluations from each automatic evaluator with the random baseline. }



\begin{figure*}[t!]
    \begin{subfigure}{0.49\textwidth}
    \centering
        \includegraphics[width=0.8\linewidth, trim={7cm 0.0cm 9cm 2.0cm}, clip]{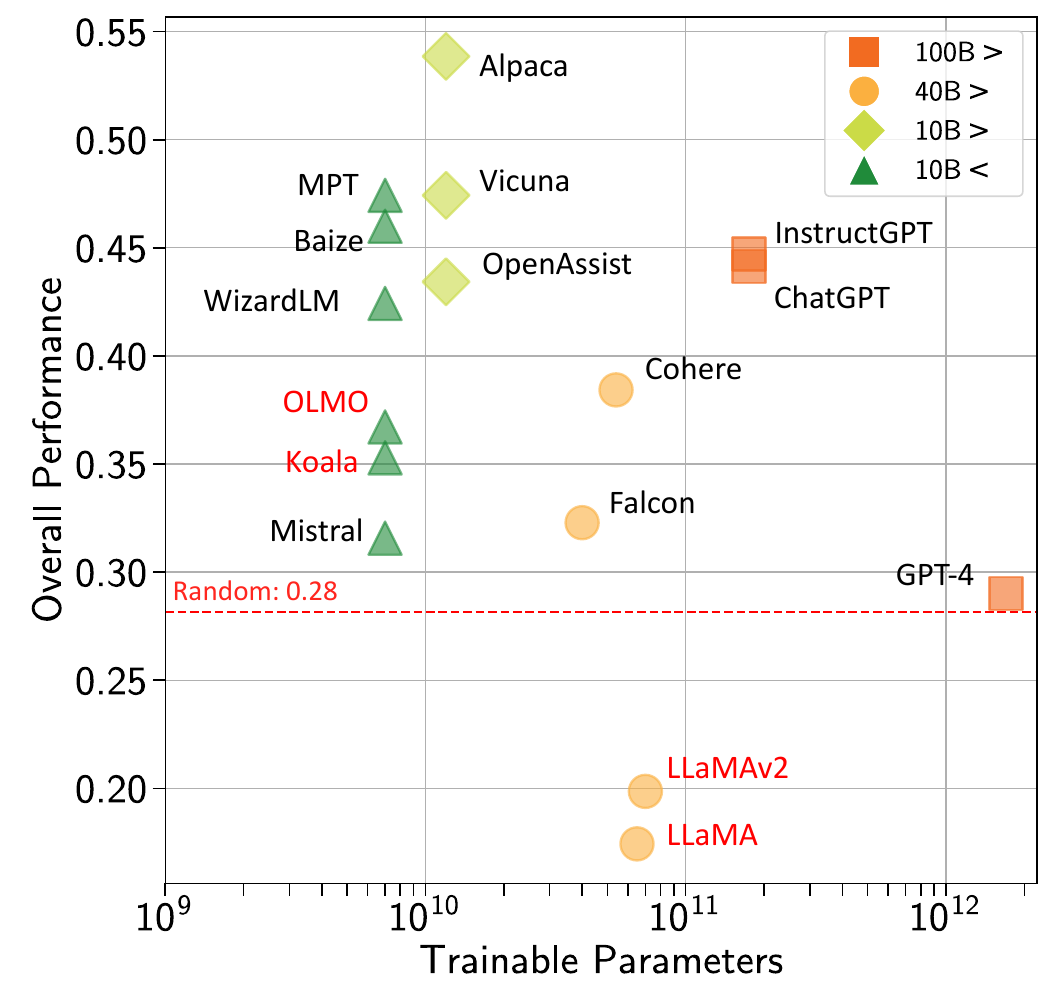}
        \caption{Average bias scores by model size}
        \label{fig:scatter}
    \end{subfigure}
    \begin{subfigure}{0.49\textwidth}
    \centering
        \includegraphics[width=0.9\linewidth, trim={7.5cm 0.0cm 6cm 0.0cm}, clip]{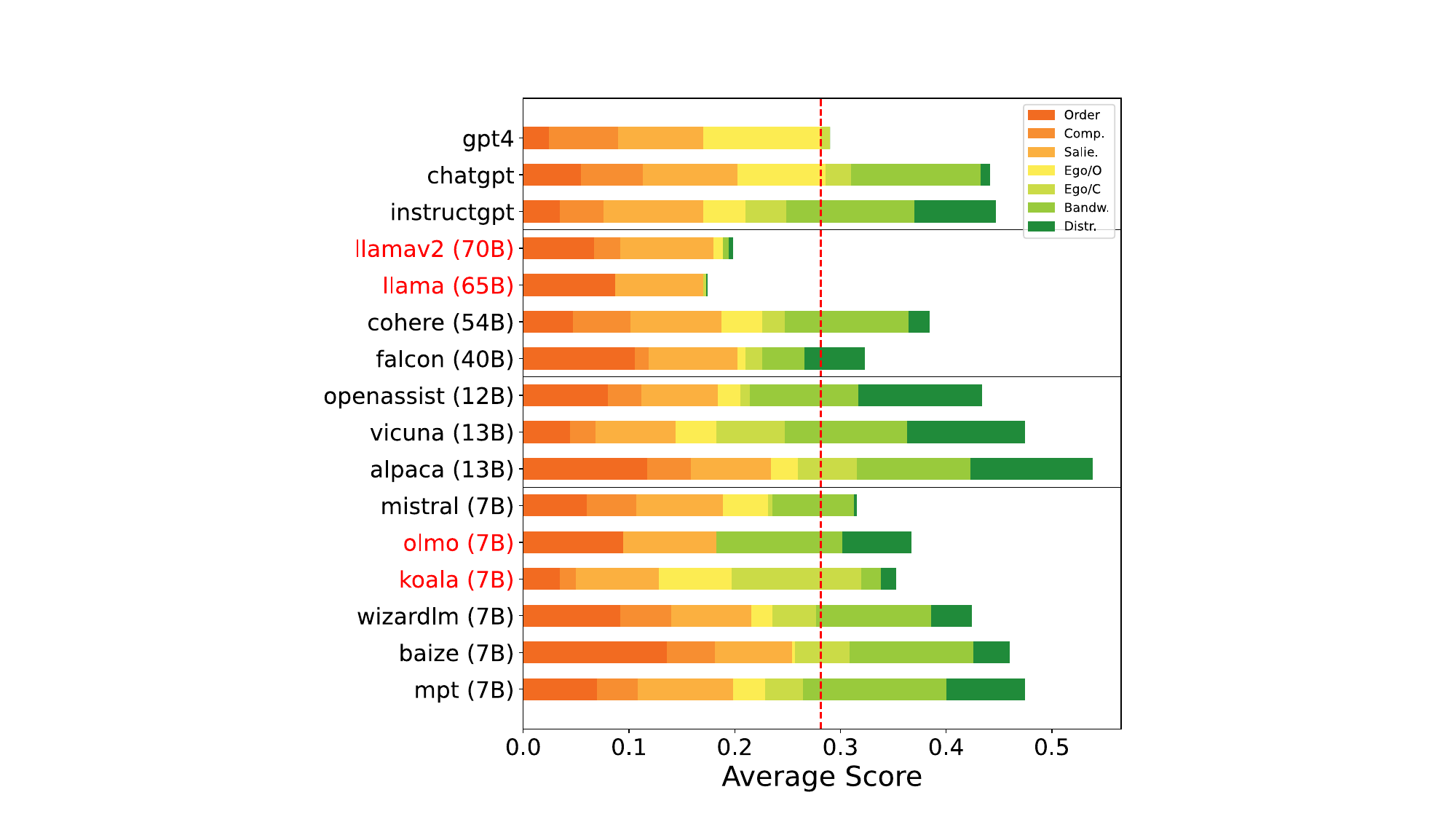}
        \caption{Breakdown of the proportional impact of each bias}
        \label{fig:barplot}
    \end{subfigure}
    \caption{Overview of major findings \new{(lower score indicates "less biased" or better performance)} of evaluator capabilities on all bias benchmarks. The red-dotted line denotes the average \textsc{Random} threshold across each bias. Models highlighted \textcolor{red}{red} indicate ones with $<80\%$ valid evaluations on 2 or more of the benchmarks.}
    \label{fig:major_breakdown}
\end{figure*}

\subsection{Human Preference Study}\label{sec:human_preference_study_setup}
We collected human preferences from six workers on Amazon mechanical Turk (AMT) platform. More details about our data collection, human annotation process, and Rank-Biased Overlap and our calculation process are presented in Appendix \ref{sec:appendix:human}.

\paragraph{Agreement between Human Preference and LLM Evaluation} 

We calculated the Rank-Biased Overlap (RBO) score \citep{webber2010similarity} to measure the \textit{agreement} between human preferences and model evaluations in ranking-generated texts across 16 different LLMs. RBO, which can vary from 0 (non-conjoint) to 1 (identical), assigns more weight to the top $k$ items in the ranked lists being compared \footnote{We concentrated 86\% of all weights on the top 5 list positions, following \citet{webber2010similarity}. }. Higher RBO score means higher agreement. Further mathematical details of RBO setup can be found in Appendix \ref{sec:appendix:human:rbo}. To properly compare machine and human preferences, we construct a ranked list for each evaluator by counting \new{each model wins}\footnote{At the time of human experiments, \textsc{LLaMA2}, \textsc{Mistral}, and \textsc{OLMo} were added later and instead involved responses by \textsc{RedPajama} and \textsc{Dolly}. Thus, the ranking of those three models was not included involving pairwise comparisons between \textbf{13} models.} from every pairwise comparison and then calculated the RBO. \new{Here, we computed the RBOs between each individual annotator and machine preferences and averaged them.}



\paragraph{Identifying Biases in Pairwise Human Preference} 

\new{To validate the gap between model judgment and humans, we conduct another study to measure the degree of bias in human evaluations as well.} We mirror the pairwise model evaluation setting from Section \ref{sec:text-evaluation-setting} for \textsc{Order Bias}, \textsc{Salience Bias}, \textsc{Bandwagon Effect}, and \textsc{Attentional Bias} for a separate human study. To obtain an effective metric, and due to the vastness of the pairwise model comparison settings, we randomly sampled 750 pairs from 25 different instructions. We then calculate the average IAA for each bias via RBO and then compute the average bias proportion across all annotators to highlight the overall influence of each bias on human judgment. 

\review{To maintain consistency with the initial study, where we used RBO as an IAA metric among human annotators for the previous N-wise ranking human experiment, we employed the same approach for the pairwise human bias experiment as opposed to Fleiss' Kappa or other pairwise agreement scores. This involved converting all pairwise rankings by humans into a ranked list of models and computing the IAA scores among the human annotators for each of the three bias experiments. However, we randomly paired models for each instruction and thus generated 750 model pairs per bias \footnote{Note that there are 25 batches in total for 750 pairs per bias and 75 human annotators}, with some models appearing either multiple times or none in those pairs. As some models may be overrepresented, we compensate for the absence of some models by applying a normalization to the rankings of appearing models across all judged pairs per human annotator. More details are described in Appendix  \ref{sec:appendix:pairwise-human-bias}, \ref{sec:appendix:pairwise-human-bias-samples}, and \ref{sec:appendix:amt-interface}.}

\begin{table*}[t!]
\centering
\small
\begin{tabularx}{0.85\linewidth}
{@{}p{2.0cm}p{0.8cm}
p{0.5cm}@{\hskip0.7cm}p{0.5cm}
p{0.5cm}@{\hskip 0.7cm}p{0.5cm}
p{0.5cm}@{\hskip 0.7cm}p{0.5cm}
@{\hskip 0.7cm}p{0.5cm}@{\hskip 0.7cm}p{0.5cm}@{\hskip 0.7cm}p{0.5cm}@{}}
\toprule
Model & Size & \multicolumn{2}{c}{\scriptsize{\textsc{Order}}} & \multicolumn{2}{c}{\scriptsize{\textsc{Comp.}}}& \multicolumn{2}{c}{\centering\scriptsize{\textsc{Egoc.}}} & \scriptsize{\textsc{Sal.}} & \scriptsize{\textsc{Band.}}  & \scriptsize{\textsc{Attn.}} \\
& & \tiny{First} & \tiny{Last} & \tiny{First} & \tiny{Last} & \tiny{Order} & \tiny{Comp.} & & & \\
\midrule
\textsc{Random} & - & 0.24 & 0.25 & 0.24 & 0.25 & 0.24 & 0.24 & 0.5 & 0.25 & 0.25 \\
\midrule
\textsc{GPT4}  & - & 0.17 & 0.06 & \cellcolor{red!21}0.46 & \cellcolor{red!8}0.33 & \cellcolor{red!54}0.78 & 0.06 & \cellcolor{red!6}0.56 & 0.0 & 0.0 \\
\textsc{ChatGPT} & 175B & \cellcolor{red!14}0.38 & 0.03 & \cellcolor{red!17}0.41 & 0.25 & \cellcolor{red!34}0.58 & 0.17 & \cellcolor{red!13}0.63 & \cellcolor{red!61}0.86 & 0.06\\
\textsc{InstructGPT} & 175B & 0.14 & 0.24 & \cellcolor{red!5}0.29 & 0.19 & \cellcolor{red!4}0.28 & \cellcolor{red!3}0.27 & \cellcolor{red!16}0.66 & \cellcolor{red!60}0.85 & \cellcolor{red!29}0.54 \\
\midrule
\textsc{LLaMAv2} & 70B & \cellcolor{red!23}0.47 & 0.08 & 0.09 & 0.17 & 0.06 & 0.0 & \cellcolor{red!12}0.62 & 0.04 & 0.03  \\
\textsc{LLaMA}  & 65B & \cellcolor{red!37}0.61 & 0.0 & 0.0 & 0.0 & 0.0&0.02 & \cellcolor{red!8}0.42 & 0.0 & 0.01 \\
\textsc{Cohere}  & 54B & \cellcolor{red!9}0.33 & 0.17 & \cellcolor{red!9}0.38 & \cellcolor{red!2}0.27 & \cellcolor{red!2}0.27 & 0.15 & \cellcolor{red!10}0.60 & \cellcolor{red!57}0.82 & 0.14 \\
\textsc{Falcon}  & 40B &  \cellcolor{red!50}0.74 & 0.03 & 0.09 & 0.18 & 0.05 & 0.11 & \cellcolor{red!9}0.59 & \cellcolor{red!3}0.28 & \cellcolor{red!15}0.40 \\
\midrule
\textsc{Alpaca}  & 13B &  0.0 & \cellcolor{red!}0.82 & 0.23 & \cellcolor{red!4}0.29 & 0.18 & \cellcolor{red!15}0.39 & \cellcolor{red!3}0.47 & \cellcolor{red!50}0.75 & \cellcolor{red!56}0.81 \\
\textsc{Vicuna}  & 13B & \cellcolor{red!8}0.32 & 0.17 & 0.17 & 0.15 & \cellcolor{red!3}0.27 & \cellcolor{red!21}0.45 & \cellcolor{red!3}0.53 & \cellcolor{red!56}0.81 & \cellcolor{red!53}0.78 \\
\textsc{OpenAssist} & 12B & \cellcolor{red!32}0.56 & 0.11 & 0.03 & 0.22 & 0.15 & 0.06 & \cellcolor{red!1}0.49 & \cellcolor{red!47}0.72 & \cellcolor{red!57}0.82 \\
\midrule
\textsc{Mistral} & 7B & \cellcolor{red!18}0.42 & 0.04 & \cellcolor{red!10}0.33 & 0.23 & \cellcolor{red!6}0.30 & 0.03 & \cellcolor{red!7}0.57 & \cellcolor{red!30}0.54 & 0.02 \\
\textsc{Olmo} & 7B & \cellcolor{red!42}0.66 & 0.0 & 0.0 & 0.0 & 0.0 & 0.0 & \cellcolor{red!12}0.38 & \cellcolor{red!50}0.83 & \cellcolor{red!22}0.46 \\
\textsc{Baize}  & 7B &  0.0 & \cellcolor{red!71}0.95 & 0.21 & \cellcolor{red!8}0.32 & 0.02 & \cellcolor{red!12}0.36 & \cellcolor{red!1}0.49 & \cellcolor{red!58}0.82 & 0.24 \\
\textsc{Koala} & 7B &  0.24 & 0.01 & 0.0 & 0.11 & \cellcolor{red!24}0.48 & \cellcolor{red!62}0.86 & \cellcolor{red!5}0.55 & 0.13 & 0.10 \\
\textsc{WizardLM} & 7B & 0.08 & \cellcolor{red!40}0.64 & 0.22 & \cellcolor{red!8}0.34 & 0.14 & \cellcolor{red!5}0.29 & \cellcolor{red!3}0.53 & \cellcolor{red!51}0.76 & \cellcolor{red!2}0.27 \\
\textsc{MPT} & 7B & \cellcolor{red!25}0.49 & 0.1 & 0.11 & \cellcolor{red!2}0.27 & 0.21 & \cellcolor{red!1}0.25 & \cellcolor{red!13}0.63 & \cellcolor{red!70}0.95 & \cellcolor{red!27}0.52 \\ 
\bottomrule
\end{tabularx}
\caption{A comparison of 16 models with different ranges of model sizes across six different bias benchmarks. 
A higher proportion indicates worse (more biased) performance. 
For \textsc{Order Bias} and \textsc{Compassion Fade}, \textit{First} indicates the proportion of responses preferring the first ordered response and \textit{Last} for the last ordered response. For \textsc{Salience Bias}, models with scores less than 0.5 prefer responses with \textbf{fewer} tokens, and scores above 0.5 prefer responses with \textbf{more} tokens. The background color of each metric is determined by the difference between the value and the corresponding \textsc{Random} metric (darker shade indicates stronger bias).}
\label{results:main}
\end{table*}

\section{Results and Discussion}

For each bias, we analyze the performance of each of the 16 models as evaluators. We provide a visual breakdown of the proportional impact of the average performance of each model as unbiased evaluators in Fig. \ref{fig:major_breakdown} based on the results relative to the \textsc{Random} baseline in Table \ref{results:main}. On average, we see that models within the $10B$ size range are most affected by each bias benchmark in Fig. \ref{fig:scatter}. Notably, we see that the implicit biases contribute similarly to each models' overall bias scores, indicating that scaling model size does not reduce implicit biases in evaluators. 




\subsection{Bias Analysis}
\paragraph{Implicit Biases}

We first examine the performance of each evaluator on the implicit bias benchmarks for \textsc{Order Bias, Compassion Fade, Salience Bias} and \textsc{Egocentric Bias}. 
For the \textsc{Order Bias} benchmark in Table \ref{results:main}, we observe that most models (11/15) tend to be drawn towards either the first- or last-ordered model in each of the pairwise comparisons. Notably, within the second size group ($>$$40B$), the first-ordered system was strongly favored in over 50\%.

For \textsc{Compassion Fade}, since it is difficult to interpret its impact by the metrics independently, we jointly compare the results with the ones from \textsc{Order Bias}. For an unbiased evaluator that is not influenced by \new{identifiable names}, we expect the results for \textsc{Compassion Fade} to be relatively similar to the \textsc{Order Bias} benchmark. However, we see in Table \ref{results:main} that all models are dramatically influenced by real model names. 
Although this phenomenon may be akin to injecting random names, the disparity between \textsc{Order} and \textsc{Compassion Fade} results support our hypothesis that recognizable names influence evaluations in contrast to anonymized ones.
\new{In addition, we also note that \textsc{Olmo} sees a drastic decrease in performance. This might be attributed to the model's inability to follow more complex instructions from its training.}

For \textsc{Egocentric Bias}, in the anonymized aliases, the largest models as well as \textsc{Koala} tend to prefer their own responses ($>50\%$) with the exception of \textsc{InstructGPT}. However, with real model names (\textsc{Compassion)}, we see a large drop in self-preference for models in the largest size group ($>$$100B$) models, but this may be attributed to a large increase in bias for each position. On average, we see an increase in self-preference with real model names amongst the two smaller size groups, notably \textsc{Koala} sees a 100\% increase in preference. 

For \textsc{Salience Bias}, we observe that the larger models in the first and second size groups are more strongly affected by longer responses, which align with findings from other works \citep{wu2023style, zheng2023judging}. However, smaller models (excluding \textsc{MPT}) tend to be less influenced by the length of the responses, suggesting that smaller models in the third and fourth size groups are less impacted in their evaluations in consideration of the \new{length of the text}. 

\new{For models such as ChatGPT, the \textsc{Egocentric bias} may be unfair because their generations are indeed better, or in \textsc{Salience}, the longer generations indeed have higher quality. For further insight in decoupling these factors, we include supplementary experiments viewed in Appendix \ref{sec:appendix:supplementary}.}

\subsubsection{Identifying Egocentric and Salience Bias}

\review{We also discuss the evaluation criteria for identifying \textsc{Egocentric} and \textsc{Salience} biases, which may be more appropriately evaluated conditioned on underlying generation quality and model size.}

\review{We select a few model representative models for clarity viewed in Table \ref{tab:supplementary:rebuttal_decoupling}. Generally, most models stay consistent with their preference for longer/shorter responses conditioned on either generation's quality, although some flip their preferences (to only a small effect however). For further insight, we compute the generation quality using reference-based metrics via \textit{BERTScore}. From this, all models produce nearly the same quality of generations with respect to the reference answer ($\sim$0.81 to 0.86 for F1), highlighting that identifying \textsc{Egocentric} or \textsc{Salience} bias is most likely not dependent on generation quality.}

\begin{table}[t]
\small
\centering
\begin{tabular}{lccc}
\toprule
Model & Salience & Salience$_{large}$ & Salience$_{small}$ \\
\midrule
\textsc{GPT4}        & 0.56 & 0.71 & 0.46 \\
\textsc{ChatGPT}     & 0.63 & 0.84 & 0.56 \\
\textsc{LlamaV2}     & 0.62 & 0.75 & 0.53 \\
\textsc{Cohere}      & 0.60 & 0.71 & 0.56 \\
\textsc{Vicuna}      & 0.53 & 0.57 & 0.51 \\
\textsc{Mistral}     & 0.57 & 0.68 & 0.50 \\
\textsc{OLMO}        & 0.38 & 0.45 & 0.29 \\
\bottomrule
\end{tabular}
\caption{\textsc{Salience} of selected models preferring generations from large models vs. small models. We see only small deviations in preference of large and small models' generations with respect to saliency.}
\label{tab:supplementary:rebuttal_decoupling}
\vspace{-4mm}
\end{table}

\paragraph{Induced Biases}
Next, we evaluate the performance of each evaluator on the induced bias benchmarks: \textsc{Bandwagon Effect} and \textsc{Attentional Bias}. For \textsc{Bandwagon Effect}, we observe that almost all models (11/15) are heavily influenced in which $>70\%$ of evaluations on average followed the bandwagon preference regardless of text quality. Although we only included a simple fake statistic (e.g. \textit{85\% of people preferred ``System Star``}), we see that evaluators can be heavily influenced by this external information. \new{To observe a correlation between the biased tendency and the percentage, we include additional results in Appendix \ref{sec:appendix:supplementary:bandwagon} }

For \textsc{Attentional Bias}, we see that around half of the models' rankings are influenced by irrelevant information. 
Specifically, we see that models in the third size group ($>$$10B$) were the most strongly impacted by the distracting information, with $>80\%$ of evaluations being counted as distracted. On the other hand, API-based models such as \textsc{ChatGPT} and \textsc{Cohere} remained robust against these distractions in their rankings. We include the list of distractions we use in Appendix \ref{sec:appendix:prompting}. 

Lastly, we address specific models such as \textsc{LLaMAv2}, \textsc{LLaMA}, \textsc{Koala}, and \textsc{OLMO} that show abnormal results on most of the benchmarks. This can be attributed to their low valid response rates, displayed in Table \ref{results:validity} in Appendix \ref{sec:appendix:supplementary}, which may be explained by our prompting format or the capabilities of the model themselves, likely as they are not instruction-tuned. \new{Although these models display lower performance when extracting evaluations, if a model is not strong enough to produce valid outputs, we assume those models are not strong enough to be used for evaluations. And as we don’t consider invalid responses within the study, we only apply our findings to ones that produced valid evaluations, in which most models exhibit cognitive biases from our benchmark. Although the correlation between valid response rates and bias can provide more insight into model capabilities, it is not within the scope of our findings.}

\begin{figure}[t!]
    \centering
    \includegraphics[width=\linewidth]{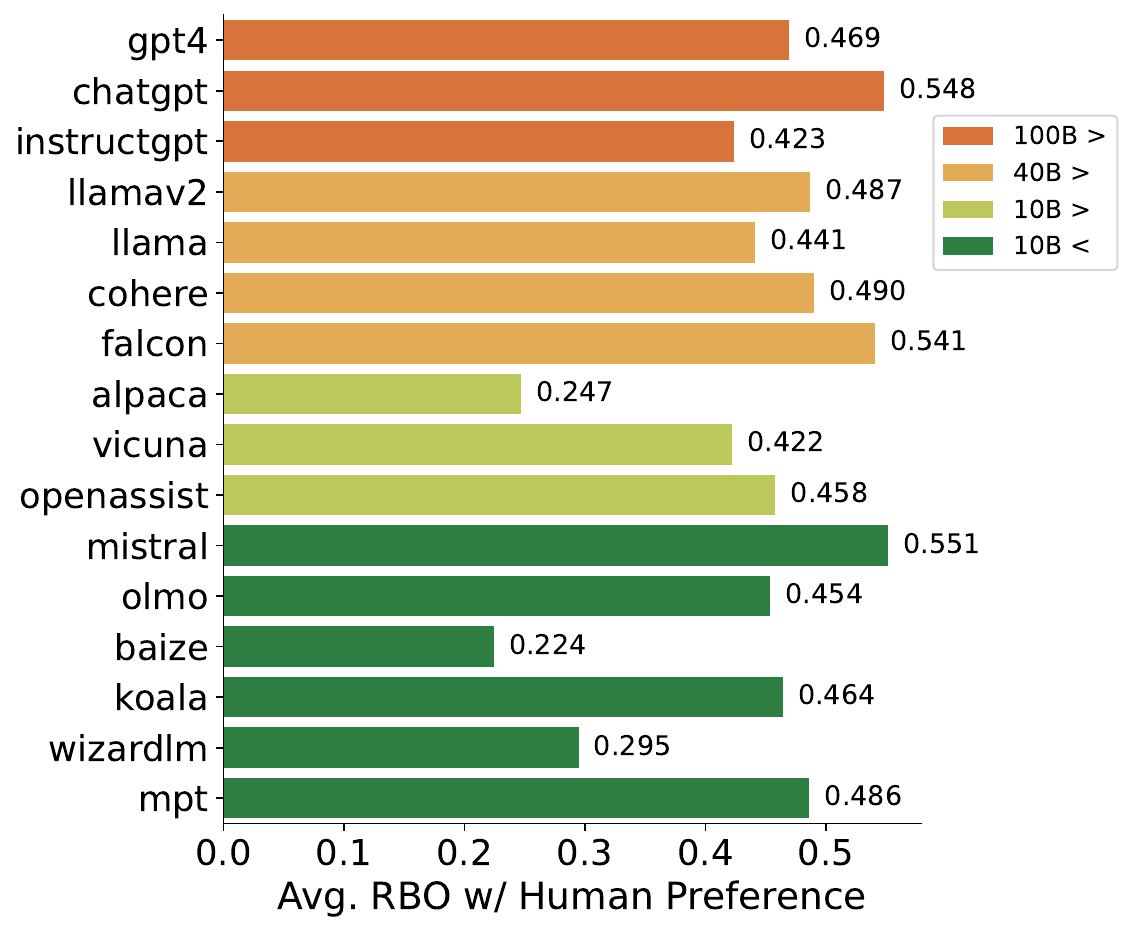}
    \caption{Correlation with human judgment. We show the average Rank-Biased Overlap (RBO) scores between aggregated human preferences and each of the 16 LLMs. Higher RBO means higher similarity.}
    \label{fig:modeltohuman}
    \vspace{-4mm}
\end{figure}

\subsection{Agreement Between Human Preferences and Model Evaluations}

\paragraph{N-rankwise Human Preference (N=13)} 
The average RBO among the six AMT workers is \textbf{\new{0.54}}, which signifies a modest but reasonable consensus among workers in ranking the LLM outputs, given the challenges of ranking all LLM-generated outputs. From this, we calculate the average RBO between human and model preferences to be \textbf{\new{0.44}}, indicating that model evaluations do not closely align with human preferences.

\new{Figure \ref{fig:modeltohuman} presents the average RBO scores \new{between a model and each of human preferences}. While \textsc{Mistral} and \textsc{ChatGPT} achieved the highest RBO scores, most of the remaining models demonstrated lower agreement with human preferences. Smaller models also tend to misalign with an overall human preference, as the average RBO of models of size greater or smaller than 10B are 0.37 and 0.41, respectively, compared to $>$$40B$ (0.49) and $>$$100B$ (0.48).}

\review{Furthermore, we present additional results on the variance of pairwise RBOs based on our annotations by six human annotators for the N=13-wise ranking experiment. Table \ref{tab:varience_spread} presents that the variance of all pairwise RBOs among humans is 0.004, indicating minimal disparity amongst all annotators. It is more clearly observed that any pairwise RBO between two annotators is higher than the average agreement between humans and models (0.44).}

\paragraph{Bias in Pairwise Human Preference} 

The average RBO scores were 0.39 (\textsc{Order Bias}), 0.50 (\textsc{Bandwagon Effect}), and 0.43 (\textsc{Attentional Bias}), indicating modest agreement \footnote{\new{Note that we considered these scores, which might initially appear low, as relatively high when considering the impact of biases that can affect individuals to varying degrees.}} amongst human annotators in a pairwise selection setting. The average proportion of biased responses across all human annotators for \textsc{Order Bias}, \textsc{Salience Bias}, \textsc{Bandwagon Effect}, and \textsc{Attentional Bias} are presented in Table \ref{table:human:bias}. Compared to humans, \textsc{Vicuna} shows higher or similar bias proportions on all of the four bias types, where its \newpage \noindent \textsc{Attentional Bias} proportion particularly exceeds humans by more than twice. 

\review{We view that humans still exhibit biases when making their preferences on pairwise LLM evaluations, but less than LLM evaluators on average. Similarly, on the induced bias benchmarks, humans were less affected by \textsc{Bandwagon effect} and \textsc{Attentional} bias, highlighting a prevalent gap between model judgment capabilities and human ones, in which human-like biases are more intensified.}

\begin{table}[t] 
\begin{center}
\small
\resizebox{0.46\textwidth}{!}{
\begin{tabular}{l|cccc}
\toprule
& \scriptsize{\textsc{Order}} & \scriptsize{\textsc{Salie.}} & \scriptsize{\textsc{Bandw.}} & \scriptsize{\textsc{Atten.}} \\
& & & & \\
\toprule
\textsc{Human} & 0.20 & 0.52 & 0.47 & 0.35 \\
\midrule
\textsc{Vicuna} & 0.32 & 0.53 & 0.81 & 0.78 \\
\bottomrule
\end{tabular}}
\end{center}
\caption{Comparison of Human bias vs. Vicuna \review{for the proportion of biased evaluations}. For \textsc{Order}, we show the worst performance.}
\label{table:human:bias}
\end{table}

\begin{table}[t]
\small
\centering
\begin{tabular}{lcccccc}
\toprule
     & A1 & A2 & A3 & A4 & A5 & A6 \\
\midrule
A1 & 1     & 0.694 & 0.466 & 0.469 & 0.511 & 0.484 \\
A2 &       & 1     & 0.471 & 0.483 & 0.515 & 0.512 \\
A3 &       &       & 1     & 0.572 & 0.589 & 0.548 \\
A4 &       &       &       & 1     & 0.607 & 0.536 \\
A5 &       &       &       &       & 1     & 0.597 \\
A6 &       &       &       &       &       & 1     \\
\bottomrule
\end{tabular}
\caption{Upper triangle agreement between each human annotator. We see the agreement between each worker is in general much higher than the agreement between human and LLMs (0.44).}
\label{tab:varience_spread}
\vspace{-4mm}
\end{table}


\section{Conclusion}

In this paper, we analyze 16 recently developed LLMs for their suitability as automatic text quality annotators in Q/A settings. We introduce a new benchmark \method to assess their evaluation performance against 1) \textbf{Implicit} and 2) \textbf{Induced} biases. Additionally, we compare LLM evaluations to human preferences and find only a \new{44}\% average agreement. Our results indicate that most LLMs exhibit cognitive biases to a greater extent than humans, suggesting that LLMs are still unsuitable as fair and reliable automatic evaluators. In the future, potential de-biasing methods provide another area of interest in reducing each bias. For example, techniques such as chain-of-thought (CoT) reasoning or other alignment methods can perhaps be employed to reduce the bias for current models. 

\section*{Limitations} We acknowledge a few limitations within our study. Some models reach very low valid response rates, which may be due to the prompting format. With model-specific prompts, we may be able to extract more clear results for each bias. Additionally, we address the fairly subpar IAA within our human judgment study. This may be due to the difficulty of the task, asking MTurk annotators to rank 15 models to limit the number of comparisons required in a pairwise format, but also increases the complexity of the task itself, which may have caused lower quality in the annotations.

\new{We also highlight the stability of our findings in the long term. As LLM research is rapidly growing, the capabilities of language models can scale exponentially with time. As such, with new developments being discovered frequently, previous LLM performance on our bias benchmarks may quickly become outdated (i.e. \textsc{InstructGPT} can be considered an "outdated LLM," as the API is also no longer offered on OpenAI's platforms).}

\section*{Acknowledgements}

This work was mainly supported by the research gift from Grammarly. We also thank Minnesota NLP group members for providing us with valuable feedback and comments on the initial draft. 

\bibliography{anthology,custom}

\begin{thebibliography}{61}
\expandafter\ifx\csname natexlab\endcsname\relax\def\natexlab#1{#1}\fi

\bibitem[{Almazrouei et~al.(2023)Almazrouei, Alobeidli, Alshamsi, Cappelli, Cojocaru, Debbah, Goffinet, Heslow, Launay, Malartic, Noune, Pannier, and Penedo}]{falcon40b}
Ebtesam Almazrouei, Hamza Alobeidli, Abdulaziz Alshamsi, Alessandro Cappelli, Ruxandra Cojocaru, Merouane Debbah, Etienne Goffinet, Daniel Heslow, Julien Launay, Quentin Malartic, Badreddine Noune, Baptiste Pannier, and Guilherme Penedo. 2023.
\newblock {Falcon-40B}: an open large language model with state-of-the-art performance.

\bibitem[{Bai et~al.(2023)Bai, Ying, Cao, Lv, He, Wang, Yu, Zeng, Xiao, Lyu, Zhang, Li, and Hou}]{bai2023benchmarking}
Yushi Bai, Jiahao Ying, Yixin Cao, Xin Lv, Yuze He, Xiaozhi Wang, Jifan Yu, Kaisheng Zeng, Yijia Xiao, Haozhe Lyu, Jiayin Zhang, Juanzi Li, and Lei Hou. 2023.
\newblock \href {http://arxiv.org/abs/2306.04181} {Benchmarking foundation models with language-model-as-an-examiner}.

\bibitem[{Beeching et~al.(2023)Beeching, Fourrier, Habib, Han, Lambert, Rajani, Sanseviero, Tunstall, and Wolf}]{open-llm-leaderboard}
Edward Beeching, Clémentine Fourrier, Nathan Habib, Sheon Han, Nathan Lambert, Nazneen Rajani, Omar Sanseviero, Lewis Tunstall, and Thomas Wolf. 2023.
\newblock Open llm leaderboard.
\newblock \url{https://huggingface.co/spaces/HuggingFaceH4/open_llm_leaderboard}.

\bibitem[{Brown et~al.(2020)Brown, Mann, Ryder, Subbiah, Kaplan, Dhariwal, Neelakantan, Shyam, Sastry, Askell, Agarwal, Herbert-Voss, Krueger, Henighan, Child, Ramesh, Ziegler, Wu, Winter, Hesse, Chen, Sigler, Litwin, Gray, Chess, Clark, Berner, McCandlish, Radford, Sutskever, and Amodei}]{brown2020language}
Tom~B. Brown, Benjamin Mann, Nick Ryder, Melanie Subbiah, Jared Kaplan, Prafulla Dhariwal, Arvind Neelakantan, Pranav Shyam, Girish Sastry, Amanda Askell, Sandhini Agarwal, Ariel Herbert-Voss, Gretchen Krueger, Tom Henighan, Rewon Child, Aditya Ramesh, Daniel~M. Ziegler, Jeffrey Wu, Clemens Winter, Christopher Hesse, Mark Chen, Eric Sigler, Mateusz Litwin, Scott Gray, Benjamin Chess, Jack Clark, Christopher Berner, Sam McCandlish, Alec Radford, Ilya Sutskever, and Dario Amodei. 2020.
\newblock \href {http://arxiv.org/abs/2005.14165} {Language models are few-shot learners}.

\bibitem[{Butts et~al.(2019)Butts, Lunt, Freling, and Gabriel}]{butts2019compassion}
Marcus~M. Butts, Devin~C. Lunt, Traci~L. Freling, and Allison~S. Gabriel. 2019.
\newblock \href {https://doi.org/https://doi.org/10.1016/j.obhdp.2018.12.006} {Helping one or helping many? a theoretical integration and meta-analytic review of the compassion fade literature}.
\newblock \emph{Organizational Behavior and Human Decision Processes}, 151:16--33.

\bibitem[{Chen et~al.(2023)Chen, Wang, Jiang, Shi, and Xu}]{chen2023exploring}
Yi~Chen, Rui Wang, Haiyun Jiang, Shuming Shi, and Ruifeng Xu. 2023.
\newblock \href {http://arxiv.org/abs/2304.00723} {Exploring the use of large language models for reference-free text quality evaluation: A preliminary empirical study}.

\bibitem[{Chiang et~al.(2023)Chiang, Li, Lin, Sheng, Wu, Zhang, Zheng, Zhuang, Zhuang, Gonzalez, Stoica, and Xing}]{vicuna2023}
Wei-Lin Chiang, Zhuohan Li, Zi~Lin, Ying Sheng, Zhanghao Wu, Hao Zhang, Lianmin Zheng, Siyuan Zhuang, Yonghao Zhuang, Joseph~E. Gonzalez, Ion Stoica, and Eric~P. Xing. 2023.
\newblock \href {https://lmsys.org/blog/2023-03-30-vicuna/} {Vicuna: An open-source chatbot impressing gpt-4 with 90\%* chatgpt quality}.

\bibitem[{Cohen et~al.(2023)Cohen, Hamri, Geva, and Globerson}]{cohen2023lm}
Roi Cohen, May Hamri, Mor Geva, and Amir Globerson. 2023.
\newblock \href {http://arxiv.org/abs/2305.13281} {Lm vs lm: Detecting factual errors via cross examination}.

\bibitem[{Conover et~al.(2023)Conover, Hayes, Mathur, Xie, Wan, Shah, Ghodsi, Wendell, Zaharia, and Xin}]{DatabricksBlog2023DollyV2}
Mike Conover, Matt Hayes, Ankit Mathur, Jianwei Xie, Jun Wan, Sam Shah, Ali Ghodsi, Patrick Wendell, Matei Zaharia, and Reynold Xin. 2023.
\newblock \href {https://www.databricks.com/blog/2023/04/12/dolly-first-open-commercially-viable-instruction-tuned-llm} {Free dolly: Introducing the world's first truly open instruction-tuned llm}.

\bibitem[{Das et~al.(2024)Das, Langis, Martin-Boyle, Kim, Lee, Kim, Hayati, Owan, Hu, Parkar, Koo, Park, Tyagi, Ferland, Roy, Liu, and Kang}]{das2024surface}
Debarati Das, Karin~De Langis, Anna Martin-Boyle, Jaehyung Kim, Minhwa Lee, Zae~Myung Kim, Shirley~Anugrah Hayati, Risako Owan, Bin Hu, Ritik Parkar, Ryan Koo, Jonginn Park, Aahan Tyagi, Libby Ferland, Sanjali Roy, Vincent Liu, and Dongyeop Kang. 2024.
\newblock \href {http://arxiv.org/abs/2401.14698} {Under the surface: Tracking the artifactuality of llm-generated data}.

\bibitem[{Dziri et~al.(2023)Dziri, Lu, Sclar, Li, Jiang, Lin, West, Bhagavatula, Bras, Hwang, Sanyal, Welleck, Ren, Ettinger, Harchaoui, and Choi}]{dziri2023faith}
Nouha Dziri, Ximing Lu, Melanie Sclar, Xiang~Lorraine Li, Liwei Jiang, Bill~Yuchen Lin, Peter West, Chandra Bhagavatula, Ronan~Le Bras, Jena~D. Hwang, Soumya Sanyal, Sean Welleck, Xiang Ren, Allyson Ettinger, Zaid Harchaoui, and Yejin Choi. 2023.
\newblock \href {http://arxiv.org/abs/2305.18654} {Faith and fate: Limits of transformers on compositionality}.

\bibitem[{Fan et~al.(2019)Fan, Jernite, Perez, Grangier, Weston, and Auli}]{eli5_lfqa}
Angela Fan, Yacine Jernite, Ethan Perez, David Grangier, Jason Weston, and Michael Auli. 2019.
\newblock \href {https://doi.org/10.18653/v1/p19-1346} {{ELI5:} long form question answering}.
\newblock In \emph{Proceedings of the 57th Conference of the Association for Computational Linguistics, {ACL} 2019, Florence, Italy, July 28- August 2, 2019, Volume 1: Long Papers}, pages 3558--3567. Association for Computational Linguistics.

\bibitem[{Fu et~al.(2023)Fu, Ng, Jiang, and Liu}]{fu2023gptscore}
Jinlan Fu, See-Kiong Ng, Zhengbao Jiang, and Pengfei Liu. 2023.
\newblock \href {http://arxiv.org/abs/2302.04166} {Gptscore: Evaluate as you desire}.

\bibitem[{Gao et~al.(2021)Gao, Tow, Biderman, Black, DiPofi, Foster, Golding, Hsu, McDonell, Muennighoff, Phang, Reynolds, Tang, Thite, Wang, Wang, and Zou}]{eval-harness}
Leo Gao, Jonathan Tow, Stella Biderman, Sid Black, Anthony DiPofi, Charles Foster, Laurence Golding, Jeffrey Hsu, Kyle McDonell, Niklas Muennighoff, Jason Phang, Laria Reynolds, Eric Tang, Anish Thite, Ben Wang, Kevin Wang, and Andy Zou. 2021.
\newblock \href {https://doi.org/10.5281/zenodo.5371628} {A framework for few-shot language model evaluation}.

\bibitem[{Gao et~al.(2023)Gao, Ruan, Sun, Yin, Yang, and Wan}]{gao2023humanlike}
Mingqi Gao, Jie Ruan, Renliang Sun, Xunjian Yin, Shiping Yang, and Xiaojun Wan. 2023.
\newblock \href {http://arxiv.org/abs/2304.02554} {Human-like summarization evaluation with chatgpt}.

\bibitem[{Gekhman et~al.(2023)Gekhman, Herzig, Aharoni, Elkind, and Szpektor}]{gekhman2023trueteacher}
Zorik Gekhman, Jonathan Herzig, Roee Aharoni, Chen Elkind, and Idan Szpektor. 2023.
\newblock \href {http://arxiv.org/abs/2305.11171} {Trueteacher: Learning factual consistency evaluation with large language models}.

\bibitem[{Geng et~al.(2023)Geng, Gudibande, Liu, Wallace, Abbeel, Levine, and Song}]{koala_blogpost_2023}
Xinyang Geng, Arnav Gudibande, Hao Liu, Eric Wallace, Pieter Abbeel, Sergey Levine, and Dawn Song. 2023.
\newblock \href {https://bair.berkeley.edu/blog/2023/04/03/koala/} {Koala: A dialogue model for academic research}.
\newblock Blog post.

\bibitem[{Geva et~al.(2021)Geva, Khashabi, Segal, Khot, Roth, and Berant}]{geva2021strategyqa}
Mor Geva, Daniel Khashabi, Elad Segal, Tushar Khot, Dan Roth, and Jonathan Berant. 2021.
\newblock {Did Aristotle Use a Laptop? A Question Answering Benchmark with Implicit Reasoning Strategies}.
\newblock \emph{Transactions of the Association for Computational Linguistics (TACL)}.

\bibitem[{Groeneveld et~al.(2024)Groeneveld, Beltagy, Walsh, Bhagia, Kinney, Tafjord, Jha, Ivison, Magnusson, Wang, Arora, Atkinson, Authur, Chandu, Cohan, Dumas, Elazar, Gu, Hessel, Khot, Merrill, Morrison, Muennighoff, Naik, Nam, Peters, Pyatkin, Ravichander, Schwenk, Shah, Smith, Strubell, Subramani, Wortsman, Dasigi, Lambert, Richardson, Zettlemoyer, Dodge, Lo, Soldaini, Smith, and Hajishirzi}]{groeneveld2024olmo}
Dirk Groeneveld, Iz~Beltagy, Pete Walsh, Akshita Bhagia, Rodney Kinney, Oyvind Tafjord, Ananya~Harsh Jha, Hamish Ivison, Ian Magnusson, Yizhong Wang, Shane Arora, David Atkinson, Russell Authur, Khyathi~Raghavi Chandu, Arman Cohan, Jennifer Dumas, Yanai Elazar, Yuling Gu, Jack Hessel, Tushar Khot, William Merrill, Jacob Morrison, Niklas Muennighoff, Aakanksha Naik, Crystal Nam, Matthew~E. Peters, Valentina Pyatkin, Abhilasha Ravichander, Dustin Schwenk, Saurabh Shah, Will Smith, Emma Strubell, Nishant Subramani, Mitchell Wortsman, Pradeep Dasigi, Nathan Lambert, Kyle Richardson, Luke Zettlemoyer, Jesse Dodge, Kyle Lo, Luca Soldaini, Noah~A. Smith, and Hannaneh Hajishirzi. 2024.
\newblock \href {http://arxiv.org/abs/2402.00838} {Olmo: Accelerating the science of language models}.

\bibitem[{Hayati et~al.(2021)Hayati, Kang, and Ungar}]{hayati-etal-2021-bert}
Shirley~Anugrah Hayati, Dongyeop Kang, and Lyle Ungar. 2021.
\newblock \href {https://doi.org/10.18653/v1/2021.emnlp-main.510} {Does {BERT} learn as humans perceive? understanding linguistic styles through lexica}.
\newblock In \emph{Proceedings of the 2021 Conference on Empirical Methods in Natural Language Processing}, pages 6323--6331, Online and Punta Cana, Dominican Republic. Association for Computational Linguistics.

\bibitem[{Hendrycks et~al.(2021)Hendrycks, Burns, Basart, Zou, Mazeika, Song, and Steinhardt}]{hendryckstest2021}
Dan Hendrycks, Collin Burns, Steven Basart, Andy Zou, Mantas Mazeika, Dawn Song, and Jacob Steinhardt. 2021.
\newblock Measuring massive multitask language understanding.
\newblock \emph{Proceedings of the International Conference on Learning Representations (ICLR)}.

\bibitem[{Jiang et~al.(2023)Jiang, Sablayrolles, Mensch, Bamford, Chaplot, de~las Casas, Bressand, Lengyel, Lample, Saulnier, Lavaud, Lachaux, Stock, Scao, Lavril, Wang, Lacroix, and Sayed}]{jiang2023mistral}
Albert~Q. Jiang, Alexandre Sablayrolles, Arthur Mensch, Chris Bamford, Devendra~Singh Chaplot, Diego de~las Casas, Florian Bressand, Gianna Lengyel, Guillaume Lample, Lucile Saulnier, Lélio~Renard Lavaud, Marie-Anne Lachaux, Pierre Stock, Teven~Le Scao, Thibaut Lavril, Thomas Wang, Timothée Lacroix, and William~El Sayed. 2023.
\newblock \href {http://arxiv.org/abs/2310.06825} {Mistral 7b}.

\bibitem[{Jones and Steinhardt(2022)}]{jones2022capturing}
Erik Jones and Jacob Steinhardt. 2022.
\newblock \href {https://openreview.net/forum?id=fcO9Cgn-X-R} {Capturing failures of large language models via human cognitive biases}.
\newblock In \emph{Advances in Neural Information Processing Systems}.

\bibitem[{J{\o}rgensen et~al.(2022)J{\o}rgensen, Caccavale, Igel, and S{\o}gaard}]{jorgensen-etal-2022-multilingual}
Rasmus J{\o}rgensen, Fiammetta Caccavale, Christian Igel, and Anders S{\o}gaard. 2022.
\newblock \href {https://doi.org/10.18653/v1/2022.blackboxnlp-1.11} {Are multilingual sentiment models equally right for the right reasons?}
\newblock In \emph{Proceedings of the Fifth BlackboxNLP Workshop on Analyzing and Interpreting Neural Networks for NLP}, pages 131--141, Abu Dhabi, United Arab Emirates (Hybrid). Association for Computational Linguistics.

\bibitem[{Jung et~al.(2019)Jung, Kang, Mentch, and Hovy}]{jung-etal-2019-earlier}
Taehee Jung, Dongyeop Kang, Lucas Mentch, and Eduard Hovy. 2019.
\newblock \href {https://doi.org/10.18653/v1/D19-1327} {Earlier isn{'}t always better: Sub-aspect analysis on corpus and system biases in summarization}.
\newblock In \emph{Proceedings of the 2019 Conference on Empirical Methods in Natural Language Processing and the 9th International Joint Conference on Natural Language Processing (EMNLP-IJCNLP)}, pages 3324--3335, Hong Kong, China. Association for Computational Linguistics.

\bibitem[{Kocmi and Federmann(2023)}]{kocmi2023large}
Tom Kocmi and Christian Federmann. 2023.
\newblock \href {http://arxiv.org/abs/2302.14520} {Large language models are state-of-the-art evaluators of translation quality}.

\bibitem[{Köpf et~al.(2023)Köpf, Kilcher, von Rütte, Anagnostidis, Tam, Stevens, Barhoum, Duc, Stanley, Nagyfi, ES, Suri, Glushkov, Dantuluri, Maguire, Schuhmann, Nguyen, and Mattick}]{köpf2023openassistant}
Andreas Köpf, Yannic Kilcher, Dimitri von Rütte, Sotiris Anagnostidis, Zhi-Rui Tam, Keith Stevens, Abdullah Barhoum, Nguyen~Minh Duc, Oliver Stanley, Richárd Nagyfi, Shahul ES, Sameer Suri, David Glushkov, Arnav Dantuluri, Andrew Maguire, Christoph Schuhmann, Huu Nguyen, and Alexander Mattick. 2023.
\newblock \href {http://arxiv.org/abs/2304.07327} {Openassistant conversations -- democratizing large language model alignment}.

\bibitem[{Li et~al.(2023{\natexlab{a}})Li, Patel, and Du}]{li2023prd}
Ruosen Li, Teerth Patel, and Xinya Du. 2023{\natexlab{a}}.
\newblock \href {http://arxiv.org/abs/2307.02762} {Prd: Peer rank and discussion improve large language model based evaluations}.

\bibitem[{Li et~al.(2023{\natexlab{b}})Li, Zhang, Dubois, Taori, Gulrajani, Guestrin, Liang, and Hashimoto}]{alpaca_eval}
Xuechen Li, Tianyi Zhang, Yann Dubois, Rohan Taori, Ishaan Gulrajani, Carlos Guestrin, Percy Liang, and Tatsunori~B. Hashimoto. 2023{\natexlab{b}}.
\newblock Alpacaeval: An automatic evaluator of instruction-following models.
\newblock \url{https://github.com/tatsu-lab/alpaca_eval}.

\bibitem[{Liang et~al.(2022)Liang, Bommasani, Lee, Tsipras, Soylu, Yasunaga, Zhang, Narayanan, Wu, Kumar, Newman, Yuan, Yan, Zhang, Cosgrove, Manning, Ré, Acosta-Navas, Hudson, Zelikman, Durmus, Ladhak, Rong, Ren, Yao, Wang, Santhanam, Orr, Zheng, Yuksekgonul, Suzgun, Kim, Guha, Chatterji, Khattab, Henderson, Huang, Chi, Xie, Santurkar, Ganguli, Hashimoto, Icard, Zhang, Chaudhary, Wang, Li, Mai, Zhang, and Koreeda}]{liang2022holistic}
Percy Liang, Rishi Bommasani, Tony Lee, Dimitris Tsipras, Dilara Soylu, Michihiro Yasunaga, Yian Zhang, Deepak Narayanan, Yuhuai Wu, Ananya Kumar, Benjamin Newman, Binhang Yuan, Bobby Yan, Ce~Zhang, Christian Cosgrove, Christopher~D. Manning, Christopher Ré, Diana Acosta-Navas, Drew~A. Hudson, Eric Zelikman, Esin Durmus, Faisal Ladhak, Frieda Rong, Hongyu Ren, Huaxiu Yao, Jue Wang, Keshav Santhanam, Laurel Orr, Lucia Zheng, Mert Yuksekgonul, Mirac Suzgun, Nathan Kim, Neel Guha, Niladri Chatterji, Omar Khattab, Peter Henderson, Qian Huang, Ryan Chi, Sang~Michael Xie, Shibani Santurkar, Surya Ganguli, Tatsunori Hashimoto, Thomas Icard, Tianyi Zhang, Vishrav Chaudhary, William Wang, Xuechen Li, Yifan Mai, Yuhui Zhang, and Yuta Koreeda. 2022.
\newblock \href {http://arxiv.org/abs/2211.09110} {Holistic evaluation of language models}.

\bibitem[{Liu et~al.(2022)Liu, Shen, Zhang, Dolan, Carin, and Chen}]{liu-etal-2022-makes}
Jiachang Liu, Dinghan Shen, Yizhe Zhang, Bill Dolan, Lawrence Carin, and Weizhu Chen. 2022.
\newblock \href {https://doi.org/10.18653/v1/2022.deelio-1.10} {What makes good in-context examples for {GPT}-3?}
\newblock In \emph{Proceedings of Deep Learning Inside Out (DeeLIO 2022): The 3rd Workshop on Knowledge Extraction and Integration for Deep Learning Architectures}, pages 100--114, Dublin, Ireland and Online. Association for Computational Linguistics.

\bibitem[{Liu et~al.(2023)Liu, Iter, Xu, Wang, Xu, and Zhu}]{liu2023geval}
Yang Liu, Dan Iter, Yichong Xu, Shuohang Wang, Ruochen Xu, and Chenguang Zhu. 2023.
\newblock \href {http://arxiv.org/abs/2303.16634} {G-eval: Nlg evaluation using gpt-4 with better human alignment}.

\bibitem[{Lu et~al.(2022)Lu, Bartolo, Moore, Riedel, and Stenetorp}]{lu-etal-2022-fantastically}
Yao Lu, Max Bartolo, Alastair Moore, Sebastian Riedel, and Pontus Stenetorp. 2022.
\newblock \href {https://doi.org/10.18653/v1/2022.acl-long.556} {Fantastically ordered prompts and where to find them: Overcoming few-shot prompt order sensitivity}.
\newblock In \emph{Proceedings of the 60th Annual Meeting of the Association for Computational Linguistics (Volume 1: Long Papers)}, pages 8086--8098, Dublin, Ireland. Association for Computational Linguistics.

\bibitem[{Luo et~al.(2023)Luo, Xie, and Ananiadou}]{luo2023chatgpt}
Zheheng Luo, Qianqian Xie, and Sophia Ananiadou. 2023.
\newblock \href {http://arxiv.org/abs/2303.15621} {Chatgpt as a factual inconsistency evaluator for text summarization}.

\bibitem[{Oh et~al.(2022)Oh, Ustun, McAuley, and Kumar}]{Oh2022RankLS}
Sejoon Oh, Berk Ustun, Julian McAuley, and Srijan Kumar. 2022.
\newblock \href {https://api.semanticscholar.org/CorpusID:251442310} {Rank list sensitivity of recommender systems to interaction perturbations}.
\newblock \emph{Proceedings of the 31st ACM International Conference on Information \& Knowledge Management}.

\bibitem[{OpenAI(2023)}]{openai2023gpt4}
OpenAI. 2023.
\newblock \href {http://arxiv.org/abs/2303.08774} {Gpt-4 technical report}.

\bibitem[{Ouyang et~al.(2022)Ouyang, Wu, Jiang, Almeida, Wainwright, Mishkin, Zhang, Agarwal, Slama, Ray, Schulman, Hilton, Kelton, Miller, Simens, Askell, Welinder, Christiano, Leike, and Lowe}]{ouyang2022training}
Long Ouyang, Jeff Wu, Xu~Jiang, Diogo Almeida, Carroll~L. Wainwright, Pamela Mishkin, Chong Zhang, Sandhini Agarwal, Katarina Slama, Alex Ray, John Schulman, Jacob Hilton, Fraser Kelton, Luke Miller, Maddie Simens, Amanda Askell, Peter Welinder, Paul Christiano, Jan Leike, and Ryan Lowe. 2022.
\newblock \href {http://arxiv.org/abs/2203.02155} {Training language models to follow instructions with human feedback}.

\bibitem[{Raheja et~al.(2023)Raheja, Kumar, Koo, and Kang}]{raheja2023coedit}
Vipul Raheja, Dhruv Kumar, Ryan Koo, and Dongyeop Kang. 2023.
\newblock \href {http://arxiv.org/abs/2305.09857} {Coedit: Text editing by task-specific instruction tuning}.

\bibitem[{Ross and Sicoly(1979)}]{ross1979egocentric}
Michael Ross and Fiore Sicoly. 1979.
\newblock \href {https://doi.org/10.1037/0022-3514.37.3.322} {Egocentric biases in availability and attribution}.
\newblock \emph{Journal of Personality and Social Psychology}, 37:322--336.

\bibitem[{Schenk(2010)}]{schenk2010salience}
Deborah Schenk. 2010.
\newblock \href {https://doi.org/10.2139/ssrn.1661322} {Exploiting the salience bias in designing taxes}.
\newblock \emph{New York University Law and Economics Working Papers}, 28.

\bibitem[{Schick et~al.(2023)Schick, Yu, Jiang, Petroni, Lewis, Izacard, You, Nalmpantis, Grave, and Riedel}]{schick2023peer}
Timo Schick, Jane~A. Yu, Zhengbao Jiang, Fabio Petroni, Patrick Lewis, Gautier Izacard, Qingfei You, Christoforos Nalmpantis, Edouard Grave, and Sebastian Riedel. 2023.
\newblock \href {https://openreview.net/forum?id=KbYevcLjnc} {{PEER}: A collaborative language model}.
\newblock In \emph{The Eleventh International Conference on Learning Representations}.

\bibitem[{Schmitt-Beck(2015)}]{rudiger2015bandwagon}
Rüdiger Schmitt-Beck. 2015.
\newblock \href {https://doi.org/10.1002/9781118541555.wbiepc015} {\emph{Bandwagon Effect}}.

\bibitem[{Shen et~al.(2023)Shen, Cheng, You, and Bing}]{shen2023large}
Chenhui Shen, Liying Cheng, Yang You, and Lidong Bing. 2023.
\newblock \href {http://arxiv.org/abs/2305.13091} {Are large language models good evaluators for abstractive summarization?}

\bibitem[{Shi et~al.(2023)Shi, Chen, Misra, Scales, Dohan, Chi, Schärli, and Zhou}]{shi2023large}
Freda Shi, Xinyun Chen, Kanishka Misra, Nathan Scales, David Dohan, Ed~Chi, Nathanael Schärli, and Denny Zhou. 2023.
\newblock \href {http://arxiv.org/abs/2302.00093} {Large language models can be easily distracted by irrelevant context}.

\bibitem[{Srivastava et~al.(2023)Srivastava, Rastogi, Rao, Shoeb, Abid, Fisch, Brown, Santoro, Gupta, and et~al.}]{srivastava2023beyond}
Aarohi Srivastava, Abhinav Rastogi, Abhishek Rao, Abu Awal~Md Shoeb, Abubakar Abid, Adam Fisch, Adam~R. Brown, Adam Santoro, Aditya Gupta, and Adri{\`a} Garriga-Alonso et~al. 2023.
\newblock \href {https://openreview.net/forum?id=uyTL5Bvosj} {Beyond the imitation game: Quantifying and extrapolating the capabilities of language models}.
\newblock \emph{Transactions on Machine Learning Research}.

\bibitem[{Sun et~al.(2023)Sun, Yan, Ma, Ren, Yin, and Ren}]{sun2023chatgpt}
Weiwei Sun, Lingyong Yan, Xinyu Ma, Pengjie Ren, Dawei Yin, and Zhaochun Ren. 2023.
\newblock \href {http://arxiv.org/abs/2304.09542} {Is chatgpt good at search? investigating large language models as re-ranking agent}.

\bibitem[{Talboy and Fuller(2023)}]{talboy2023challenging}
Alaina~N. Talboy and Elizabeth Fuller. 2023.
\newblock \href {http://arxiv.org/abs/2304.01358} {Challenging the appearance of machine intelligence: Cognitive bias in llms and best practices for adoption}.

\bibitem[{Taori et~al.(2023)Taori, Gulrajani, Zhang, Dubois, Li, Guestrin, Liang, and Hashimoto}]{alpaca}
Rohan Taori, Ishaan Gulrajani, Tianyi Zhang, Yann Dubois, Xuechen Li, Carlos Guestrin, Percy Liang, and Tatsunori~B. Hashimoto. 2023.
\newblock Stanford alpaca: An instruction-following llama model.
\newblock \url{https://github.com/tatsu-lab/stanford_alpaca}.

\bibitem[{Team(2023)}]{MosaicML2023Introducing}
MosaicML~NLP Team. 2023.
\newblock \href {www.mosaicml.com/blog/mpt-7b} {Introducing mpt-7b: A new standard for open-source, commercially usable llms}.
\newblock Accessed: 2023-05-05.

\bibitem[{Touvron et~al.(2023)Touvron, Martin, Stone, Albert, Almahairi, Babaei, Bashlykov, Batra, Bhargava, Bhosale, Bikel, Blecher, Ferrer, Chen, Cucurull, Esiobu, Fernandes, Fu, Fu, Fuller, Gao, Goswami, Goyal, Hartshorn, Hosseini, Hou, Inan, Kardas, Kerkez, Khabsa, Kloumann, Korenev, Koura, Lachaux, Lavril, Lee, Liskovich, Lu, Mao, Martinet, Mihaylov, Mishra, Molybog, Nie, Poulton, Reizenstein, Rungta, Saladi, Schelten, Silva, Smith, Subramanian, Tan, Tang, Taylor, Williams, Kuan, Xu, Yan, Zarov, Zhang, Fan, Kambadur, Narang, Rodriguez, Stojnic, Edunov, and Scialom}]{touvron2023llama2}
Hugo Touvron, Louis Martin, Kevin Stone, Peter Albert, Amjad Almahairi, Yasmine Babaei, Nikolay Bashlykov, Soumya Batra, Prajjwal Bhargava, Shruti Bhosale, Dan Bikel, Lukas Blecher, Cristian~Canton Ferrer, Moya Chen, Guillem Cucurull, David Esiobu, Jude Fernandes, Jeremy Fu, Wenyin Fu, Brian Fuller, Cynthia Gao, Vedanuj Goswami, Naman Goyal, Anthony Hartshorn, Saghar Hosseini, Rui Hou, Hakan Inan, Marcin Kardas, Viktor Kerkez, Madian Khabsa, Isabel Kloumann, Artem Korenev, Punit~Singh Koura, Marie-Anne Lachaux, Thibaut Lavril, Jenya Lee, Diana Liskovich, Yinghai Lu, Yuning Mao, Xavier Martinet, Todor Mihaylov, Pushkar Mishra, Igor Molybog, Yixin Nie, Andrew Poulton, Jeremy Reizenstein, Rashi Rungta, Kalyan Saladi, Alan Schelten, Ruan Silva, Eric~Michael Smith, Ranjan Subramanian, Xiaoqing~Ellen Tan, Binh Tang, Ross Taylor, Adina Williams, Jian~Xiang Kuan, Puxin Xu, Zheng Yan, Iliyan Zarov, Yuchen Zhang, Angela Fan, Melanie Kambadur, Sharan Narang, Aurelien Rodriguez, Robert Stojnic, Sergey Edunov, and Thomas
  Scialom. 2023.
\newblock \href {http://arxiv.org/abs/2307.09288} {Llama 2: Open foundation and fine-tuned chat models}.

\bibitem[{Västfjäll et~al.(2014)Västfjäll, Slovic, Mayorga, and Peters}]{vastfjall2014}
Daniel Västfjäll, Paul Slovic, Marcus Mayorga, and Ellen Peters. 2014.
\newblock \href {https://doi.org/10.1371/journal.pone.0100115} {Compassion fade: Affect and charity are greatest for a single child in need}.
\newblock \emph{PLOS ONE}, 9(6):1--10.

\bibitem[{Wang et~al.(2020)Wang, Pruksachatkun, Nangia, Singh, Michael, Hill, Levy, and Bowman}]{wang2020superglue}
Alex Wang, Yada Pruksachatkun, Nikita Nangia, Amanpreet Singh, Julian Michael, Felix Hill, Omer Levy, and Samuel~R. Bowman. 2020.
\newblock \href {http://arxiv.org/abs/1905.00537} {Superglue: A stickier benchmark for general-purpose language understanding systems}.

\bibitem[{Wang et~al.(2023{\natexlab{a}})Wang, Liang, Meng, Sun, Shi, Li, Xu, Qu, and Zhou}]{wang2023chatgpt}
Jiaan Wang, Yunlong Liang, Fandong Meng, Zengkui Sun, Haoxiang Shi, Zhixu Li, Jinan Xu, Jianfeng Qu, and Jie Zhou. 2023{\natexlab{a}}.
\newblock \href {http://arxiv.org/abs/2303.04048} {Is chatgpt a good nlg evaluator? a preliminary study}.

\bibitem[{Wang et~al.(2023{\natexlab{b}})Wang, Li, Chen, Cai, Zhu, Lin, Cao, Liu, Liu, and Sui}]{wang2023large}
Peiyi Wang, Lei Li, Liang Chen, Zefan Cai, Dawei Zhu, Binghuai Lin, Yunbo Cao, Qi~Liu, Tianyu Liu, and Zhifang Sui. 2023{\natexlab{b}}.
\newblock \href {http://arxiv.org/abs/2305.17926} {Large language models are not fair evaluators}.

\bibitem[{Webber et~al.(2010)Webber, Moffat, and Zobel}]{webber2010similarity}
William Webber, Alistair Moffat, and Justin Zobel. 2010.
\newblock A similarity measure for indefinite rankings.
\newblock \emph{ACM Transactions on Information Systems (TOIS)}, 28(4):1--38.

\bibitem[{Wu and Aji(2023)}]{wu2023style}
Minghao Wu and Alham~Fikri Aji. 2023.
\newblock \href {http://arxiv.org/abs/2307.03025} {Style over substance: Evaluation biases for large language models}.

\bibitem[{Xu et~al.(2023{\natexlab{a}})Xu, Sun, Zheng, Geng, Zhao, Feng, Tao, and Jiang}]{xu2023wizardlm}
Can Xu, Qingfeng Sun, Kai Zheng, Xiubo Geng, Pu~Zhao, Jiazhan Feng, Chongyang Tao, and Daxin Jiang. 2023{\natexlab{a}}.
\newblock \href {http://arxiv.org/abs/2304.12244} {Wizardlm: Empowering large language models to follow complex instructions}.

\bibitem[{Xu et~al.(2023{\natexlab{b}})Xu, Guo, Duan, and McAuley}]{xu2023baize}
Canwen Xu, Daya Guo, Nan Duan, and Julian McAuley. 2023{\natexlab{b}}.
\newblock \href {http://arxiv.org/abs/2304.01196} {Baize: An open-source chat model with parameter-efficient tuning on self-chat data}.

\bibitem[{Zhao et~al.(2021)Zhao, Wallace, Feng, Klein, and Singh}]{pmlr-v139-zhao21c}
Zihao Zhao, Eric Wallace, Shi Feng, Dan Klein, and Sameer Singh. 2021.
\newblock \href {https://proceedings.mlr.press/v139/zhao21c.html} {Calibrate before use: Improving few-shot performance of language models}.
\newblock In \emph{Proceedings of the 38th International Conference on Machine Learning}, volume 139 of \emph{Proceedings of Machine Learning Research}, pages 12697--12706. PMLR.

\bibitem[{Zheng et~al.(2023)Zheng, Chiang, Sheng, Zhuang, Wu, Zhuang, Lin, Li, Li, Xing, Zhang, Gonzalez, and Stoica}]{zheng2023judging}
Lianmin Zheng, Wei-Lin Chiang, Ying Sheng, Siyuan Zhuang, Zhanghao Wu, Yonghao Zhuang, Zi~Lin, Zhuohan Li, Dacheng Li, Eric.~P Xing, Hao Zhang, Joseph~E. Gonzalez, and Ion Stoica. 2023.
\newblock \href {http://arxiv.org/abs/2306.05685} {Judging llm-as-a-judge with mt-bench and chatbot arena}.

\bibitem[{Zhuo(2023)}]{zhuo2023large}
Terry~Yue Zhuo. 2023.
\newblock \href {http://arxiv.org/abs/2304.14317} {Large language models are state-of-the-art evaluators of code generation}.

\end{thebibliography}
\bibliographystyle{acl_natbib}

\clearpage
\appendix

\section{Experimental Setup}\label{sec:appendix:experiment}

\subsection{Model Hyperparameters}

We set the same hyperparameters across models for each evaluation generation and response generation for consistency across all of the models. We limit the max new tokens generated to 128 tokens and set the temperature to 1.0. For Huggingface models, we set a repetition penalty of 1.2 and set the number of beams to 3. 

\subsection{Experimental Settings}

For models that are supported (ChatGPT, InstructGPT, GPT-4, Vicuna), we utilize Microsoft Guidance to better control LLM generations. Otherwise, we utilize the transformer pipeline library from Hugginface to retrieve each evaluation generation. Regardless of whether a models generation was collected from guidance or using the transformer pipeline, all parameters were the same. Model generation times for response generation ranged from 1 to 8 hours, and for evaluation generations ranged from 3 to 24 hours for each bias benchmark. All experiments were run on either A5000 or A6000 GPUs for models under 40B parameters. For models over 40B, A100 GPUs were utilized if an API service was not available (e.g. OpenAI, Cohere). 

\subsection{Datasets}
\textbf{Eli5} \citep{eli5_lfqa} is a long-form question-answering dataset constructed from 270$k$ threads from the ``Explain Like I'm Five'' Reddit forum. 
The online forum consists of a community for individuals to ask various questions, and answers are provided in a format that is comprehensible to five-year-olds, along with assigned scores based on community votes. 
For our purposes, we only utilize the questions and their highest-rated answers to generate responses and benchmark automatic evaluators for text-generation quality. 

\textbf{BigBench} \citep{srivastava2023beyond} is a collection of benchmarks that look to probe the abilities of language models over a diverse range of tasks. We specifically utilize the \textit{strategyQA} \citep{geva2021strategyqa} dataset, which was constructed by crowdsourcing questions from writers as well as their responses with short justifications. 
We choose the \textit{strategyQA} dataset to generate responses that require multi-step reasoning to effectively benchmark the ability of models to comprehend and compare the quality between two different explanations.

\section{Supplementary Results}\label{sec:appendix:supplementary}

\subsection{Correlation between \textsc{Bandwagon} and Percentage}\label{sec:appendix:supplementary:bandwagon}

\new{In an additional experiment, we show a modified statistic for the biased model: \texttt{"$0\%$ of people prefer \{model\}}.” If bias tendency were indeed correlated with the statistic, we would expect the evaluator model to have 0 preference for bandwagon response. Due to limited computation resources and time, we ran the additional experiments for two representative models at each size range (+ all API-based models) and presented the results below in Table \ref{tab:supplementary:bandwagon}.} 

\new{Here, one can observe that the preference choices for the bandwagon statistic greatly change (besides \textsc{GPT4} and \textsc{Vicuna}) which suggests that indeed the biased tendency is correlated with the bandwagon statistic. However, we see that \textsc{Vicuna}, in particular, is not greatly affected by the statistics. This suggests that within the prompt, the model only focuses on the phrase \texttt{“people believe that \{model\} is better”} instead of the statistic. Similarly, this may be the case for Alpaca and InstructGPT as well. We also present the results of the bandwagon test by randomly choosing a percentage between $50\%$ and $85\%$ in Table \ref{tab:supplementary:bandwagon2}. We continue see that most models demonstrate biased tendencies. }

\begin{table*}[htbp]
  \small
  \centering
    \begin{tabular}{lcccccccc}
    \toprule
    Models & \textsc{GPT-4} & \textsc{ChatGPT} & \textsc{InstuctGPT} & \textsc{Cohere} & \textsc{Alpaca} & \textsc{Vicuna} & \textsc{Baize} & \textsc{WizardLM} \\
    \midrule
    \textsc{Bandwagon (85\%)} & 0.0   & 0.86  & 0.85  & 0.82  & 0.75  & 0.81  & 0.82  & 0.76 \\
    \textsc{Bandwagon (0\%)} & 0.0   & 0.0   & 0.56  & 0.0   & 0.52  & 0.79  & 0.32  & 0.27 \\
    \bottomrule
    \end{tabular}%
    \caption{\textsc{Bandwagon} test showing a fake statistic stating $0\%$ of people prefer the chosen response.}
    \label{tab:supplementary:bandwagon}
\end{table*}%

\begin{table*}[htbp]
  \small
  \centering
    \begin{tabular}{lcccccccc}
    \toprule
    Models & \textsc{GPT-4} & \textsc{ChatGPT} & \textsc{InstuctGPT} & \textsc{Cohere} & \textsc{Alpaca} & \textsc{Vicuna} & \textsc{Baize} & \textsc{WizardLM} \\
    \midrule
    \textsc{Bandwagon (85\%)} & $0.0$  & $0.86$ & $0.85$  & $0.82$ & $0.75$ & $0.81$ & $0.82$ & $0.76$ \\
    \textsc{Bandwagon (50-85\%)} & $0.06$ & $0.70$ & $0.84$ & $0.65$ & $0.68$ & $0.96$ & $0.75$ & $0.76$ \\
    \bottomrule
    \end{tabular}%
    \caption{\textsc{Bandwagon} test showing a fake statistic stating (randomly) between $50-80\%$ of people prefer the chosen response.}
    \label{tab:supplementary:bandwagon2}
\end{table*}%

\subsection{Diverse Prompts}\label{sec:appendix:supplementary:diverse}

\new{We additionally ask each evaluator to analyze generation quality along several different aspects such as \texttt{“coherence, accuracy, factuality, and helpfulness”} following \cite{bai2023benchmarking} and \cite{zheng2023judging}. As opposed to our single-aspect format in the main section, we conjecture that these cognitive biases remain regardless of evaluation aspects. To validate this, we constructed an extended prompt viewable in \ref{sec:appendix:prompt:diverse} that incorporates different dimensions of evaluation criteria into our pairwise evaluation prompt and reported their results in Table \ref{tab:supplementary:diverse} on the \textsc{Order} benchmark. We see that by including diverse perspectives in the evaluation setting, some metrics become more pronounced (i.e. \textsc{Cohere} for \textsc{Egocentric}) or bias decreases (i.e. \textsc{Vicuna} for \textsc{Egocentric}). However, we see that the proportion of biased evaluations stays relatively consistent for most models on all benchmarks. Hence, our findings remain that models still show a large skewness in bias tendency as evaluators.} 

\begin{table*}[htbp]
\small
  \centering
  \begin{tabular}{lcccccccc}
    \toprule
    \textsc{Models} & \textsc{GPT-4} & \textsc{ChatGPT} & \textsc{InstuctGPT} & \textsc{Cohere} & \textsc{Alpaca} & \textsc{Vicuna} & \textsc{Baize} & \textsc{WizardLM} \\
    \midrule
    \textsc{Order (coh.)}  & $0.17_F$ & $0.38_F$ & $0.24_L$ & $0.33_F$ & $0.82_L$ & $0.32_F$ & $0.95_L$ & $0.64_L$ \\
    \textsc{Order (div.)} & $0.14_F$ & $0.45_F$ & $0.22_L$ & $0.23_L$ & $0.76_L$ & $0.52_F$ & $0.83_L$ & $0.68_L$ \\
    \midrule
    \textsc{Egocent. (coh.)}   & $0.78$ & $0.58$ & $0.28$ & $0.27$ & $0.18$ & $0.27$ & $0.02$ & $0.14$ \\
    \textsc{Egocent. (div.)}  & $0.80$ & $0.54$ & $0.29$ & $0.41$ & $0.18$ & $0.18$ & $0.04$ & $0.09$ \\
    \midrule 
    \textsc{Salience (coh.)}    & $0.56$ & $0.63$ & $0.66$ & $0.60$ & $0.47$ & $0.53$ & $0.49$ & $0.53$ \\
    \textsc{Salience (div.)}   & $0.57$ & $0.69$ & $0.70$ & $0.65$ & $0.49$ & $0.59$ & $0.50$ & $0.52$ \\
    \bottomrule
  \end{tabular}
  \caption{Comparison on the \textsc{Order} benchmark considering diverse evaluation perspectives. For visual clarity, we only display the bias ratio with the highest proportion and denote with subscript $x_F$ or $x_L$ 
  for first- or last-ordered bias, respectively.}
\label{tab:supplementary:diverse}
\end{table*}%

\subsection{Prompting with Ties}\label{sec:appendix:supplementary:tie}

\new{We present a modified version of the prompt in \ref{sec:appendix:supplementary:tie} that considers ties in each pairwise preference. Note that for \textsc{Salience}, if a pairwise sample was labeled as “Tie,” we do not consider it for length bias. From Table \ref{tab:supplementary:tie} we see that the inclusion of the tie option does view a considerable change in the bias benchmarks. Notably, the strongest and smallest models (\textsc{GPT-4, ChatGPT, Baize, WizardLM}) do not exhibit any change. However, we see that the mid-range models (\textsc{Alpaca, Vicuna}) and \textsc{InstructGPT} display a large preference for assigning the tie label ($\geq \sim 90 \%$) that does not present any valid results, to which we had originally only prompted two options for each evaluator to avoid this issue. The only model that demonstrated an improvement from previous bias behavior was \textsc{Cohere}.}

\begin{table*}[htbp]
\small
  \centering
  \begin{tabular}{lcccccccc}
    \toprule
    \textsc{Models} & \textsc{GPT-4} & \textsc{ChatGPT} & \textsc{InstuctGPT} & \textsc{Cohere} & \textsc{Alpaca} & \textsc{Vicuna} & \textsc{Baize} & \textsc{WizardLM} \\
    \midrule
    \textsc{Order} & $0.17_F$ & $0.38_F$ & $0.24_L$ & $0.33_F$ & $0.82_L$ & $0.32_F$ & $0.95_L$ & $0.64_L$ \\
    \textsc{Order (tie)} & $0.15_F$ & $0.43_F$ &  $0.0$ &  $0.08_L$ & $0.0$ & $0.0$ & $0.81_L$ & $0.47_L$ \\
    \textsc{Tie (\%)} & $0.01$ & $0.0$ & $0.88$ & $0.33$ & $0.95$ & $0.99$ & $0.0$ & $0.04$ \\
    \midrule
    \textsc{Egocentric}   & $0.78$ & $0.58$ & $0.28$ & $0.27$ & $0.18$ & $0.27$ & $0.02$ & $0.14$ \\
    \textsc{Egocentric (tie)} & $0.77$ & $0.60$ & $0.04$ & $0.25$ & $0.02$ & $0.0$ & $0.08$ & $0.16$ \\
    \midrule
    \textsc{Salience}    & $0.56$ & $0.63$ & $0.66$ & $0.60$ & $0.47$ & $0.53$ & $0.49$ & $0.53$ \\
    \textsc{Salience (tie)}   & $0.55$ & $0.67$ & $0.06$ & $0.35$ & $0.01$ & $0.0$ & $0.50$ & $0.48$ \\
    \bottomrule
  \end{tabular}%
  \caption{Comparison on the \textsc{Order} benchmark considering  ties. For visual clarity, we only display the bias ratio with the highest proportion and denote with subscript $x_F$ or $x_L$ 
  for first- or last-ordered bias, respectively.}
  \label{tab:supplementary:tie}%
\end{table*}%

\subsection{Decoupling Confounding Factors}\label{sec:appendix:supplementary:confounding}

\new{We particularly focus on decoupling \textsc{Egocentric} and \text{Salience}, which are the most prone to having large correlations with each other (i.e. longer generations may indeed have overall higher quality generated by much stronger models).We highlight two important aspects regarding the identification of these biases:
\begin{itemize}[leftmargin=5mm]
    \item If multiple models have a large proportion of evaluations preferring their own responses (as the evaluated pool of pairwise instances is the same for each evaluator), we reason that this may suggest “egocentric” qualities within involved evaluators, regardless of the objective strength of the models. Moreover, we see this effect is especially demonstrated between the more powerful models as well (\textsc{GPT4} \& \textsc{ChatGPT}) that suggest the presence of \textsc{Egocentric} evaluations from their disagreement.
    \item We employ various strategies to mitigate these confounding variables and isolate each analysis as much as possible. For example, we employ a “hierarchical” rubric, where some biases take priority in an evaluation. Specifically, if an evaluation shows signs of order bias by choosing A in (A first, then B) and B in (B first, then A), we do not evaluate it for \textsc{Salience} or \textsc{Egocentric} bias. 
\end{itemize} To get further insight into decoupling them, we examine additional statistics in Table \ref{tab:supplementary:confounding} displaying the proportion of \textsc{Egocentric} samples where the model’s generation was longer/shorter than the other generation. In particular, since \textsc{Olmo} only won once, and \textsc{LLaMA} never won, their \textsc{Egocentric} ratios look weird. Otherwise,
we view overall that most models (9/16) exhibit a self-preference for their own generations often when their own generations exhibit longer token length.}

\new{As above, we see that \textsc{Salience} may be associated with higher quality generations, as we see that the strongest models (GPT4, ChatGPT) often prefer their own responses when their generations are longer. Nevertheless, even in smaller models (e.g., Cohere, Koala), preference for their own generations occurs more often when they are longer. However, as we previously emphasized, if multiple models observe a self-preference for their own generations, it is difficult to associate with \textsc{Salience} as there is disagreement that is indicative of an \textsc{Egocentric} bias.}

\newpage
\subsection{Significance of Results}\label{sec:appendix:supplementary:significance}
\begin{table*}[h]
\small
\centering
\begin{tabular}{lcccc}
\toprule
\textsc{Model} & \textsc{First Order Z-Score} & \textsc{First Order P-Value} & \textsc{Last Order Z-Score} & \textsc{Last Order P-Value} \\
\midrule
\textsc{GPT-4}         & $8.45$                & $2.82e{-17}$            & $26.55$              & $2.65e{-155}$          \\
\textsc{ChatGPT}      & $-15.61$              & $6.68e{-55}$            & $32.43$              & $9.50e{-231}$          \\
\textsc{InstructGPT}  & $13.04$               & $7.08e{-39}$            & $1.17$               & $2.4e{-1}$               \\
\midrule
\textsc{LLaMAv2}      & $-33.12$              & $1.30e{-240}$           & $17.97$              & $3.53e{-72}$           \\
\textsc{LLaMA}        & $-40.84$              & $0$                 & $15.36$              & $2.95e{-53}$           \\
\textsc{Cohere}       & $-10.02$              & $1.20e{-23}$            & $8.98$               & $2.59e{-19}$           \\
\textsc{Falcon}       & $-57.36$              & $0$                 & $25.61$              & $1.24e{-144}$          \\
\midrule
\textsc{Alpaca}       & $30.67$               & $1.29e{-206}$           & $-61.45$             & $0$                \\
\textsc{Vicuna}       & $-12.49$              & $8.44e{-36}$            & $7.63$               & $2.29e{-14}$           \\
\textsc{OpenAssist}   & $-37.27$              & $4.84e{-304}$           & $13.93$              & $3.92e{-44}$           \\
\midrule
\textsc{Mistral}      & $-20.09$              & $9.13e{-90}$            & $32.56$              & $1.63e{-232}$          \\
\textsc{OLMO}         & $-54.54$              & $0$                 & $22.08$              & $4.47e{-108}$          \\
\textsc{Baize}       & $35.46$               & $1.99e{-275}$           & $-71.53$             & $0$                \\
\textsc{Koala}        & $-21.60$              & $1.77e{-103}$           & $12.18$              & $4.15e{-34}$           \\
\textsc{WizardLM}    & $20.65$               & $9.93e{-95}$            & $-40.48$             & $0$                \\
\textsc{MPT}          & $-30.71$              & $4.81e{-207}$           & $16.49$              & $4.62e{-61}$           \\
\bottomrule
\end{tabular}
\caption{Significance test scores for \textsc{Order} bias (first and last order preference) for each evaluator compared to the random baseline. We see almost every model shows significant results with $\alpha = 0.05$.}
\label{tab:supplementary:zscore_pvalue}
\end{table*}

\review{We adapt two statistical hypothesis tests based on the random bias threshold for the Order bias (first-order and last-order) benchmark in Table \ref{tab:supplementary:zscore_pvalue}.  Since we have binary outputs (bias, not biased), we conduct a two-sample Z test of proportions to determine the significance of each proportion of biased evaluations from each automatic evaluator with the random baseline. We conduct the test with the null hypothesis defined to be that “evaluator X is just as likely to make the mistake of flipping its preference according to the order of the response to the first-order as the random baseline” or equivalently:
\begin{center}
    $H_0$: the mean of Evaluator $X$ for first-order bias is not any different from random selection. 
\end{center}}
\review{On almost all of the \textsc{Order} benchmarks, the proportions of biased evaluations are statistically significant from ones by the random score. Notably, the p-values are critically small (z-scores are blown up) due to our large sample size. Also, we note that the p-value is actually not statistically significant for last-order bias in InstructGPT; however, the first-order proportions are statistically significant, indicating that one must consider the test for both positions to get the full picture of the evaluator's tendencies in reference to the random baseline. For example, if both first-order and last-order were not statistically significant from the random proportions, we might find that the evaluator is “unbiased,” but the following may also undermine the capabilities of the automatic evaluator reduced to just random choice.}



\begin{table*}[h]
  \centering  
  \begin{subtable}{\textwidth}
    \centering
    \begin{tabular}{lccc}
    \toprule
          & \textsc{GPT4}  & \textsc{ChatGPT} & \textsc{InstructGPT} \\
    \midrule
    Ego & 0.78  & 0.58  & 0.28 \\
    Longer Ego & 0.64  & 0.75  & 0.43 \\
    Shorter Ego & 0.36  & 0.25  & 0.56 \\
    \bottomrule
    \end{tabular}%
    \caption{Model Performance Comparison (\textgreater 175B)}
    \label{subtab:175b}%
  \end{subtable}%
  
  \vspace{0.5cm}
  
  \begin{subtable}{\textwidth}
    \centering
    \begin{tabular}{lcccc}
    \toprule
          & \textsc{LLaMAv2} & \textsc{LLaMA} & \textsc{Cohere} & 
        \textsc{Falcon} \\
    \midrule
    Ego & 0.06  & 0.0   & 0.27  & 0.05 \\
    Longer Ego & 0.18  & 0.0   & 0.68  & 0.6 \\
    Shorter Ego & 0.81  & 0.0   & 0.32  & 0.4 \\
    \bottomrule
    \end{tabular}%
    \caption{Model Performance Comparison (\textgreater 40B)}
    \label{subtab:40b}%
  \end{subtable}%

  \vspace{0.5cm}
  
  \begin{subtable}{\textwidth}
    \centering
    \begin{tabular}{lccc}
    \toprule
          & \textsc{Alpaca} & \textsc{Vicuna} & \textsc{OpenAssist} \\
    \midrule
    Ego & 0.18  & 0.27  & 0.15 \\
    Longer Ego & 0.38  & 0.4   & 0.71 \\
    Shorter Ego & 0.62  & 0.59  & 0.29 \\
    \bottomrule
    \end{tabular}%
    \caption{Model Performance Comparison (\textgreater 10B)}
    \label{subtab:10b}%
  \end{subtable}%

  \vspace{0.5cm}
  
  \begin{subtable}{\textwidth}
    \centering
    \begin{tabular}{lcccccc}
    \toprule
          & \textsc{Mistral} & \textsc{OLMO}  & \textsc{Baize} & \textsc{Koala} & \textsc{WizardLM} & \textsc{MPT} \\
    \midrule
    Ego & 0.3   & 0.0   & 0.02  & 0.48  & 0.14  & 0.21 \\
    Longer Ego & 0.64  & 0.0   & 0.0   & 0.56  & 0.54  & 0.83 \\
    Shorter Ego & 0.36  & 1.0   & 0.0   & 0.44  & 0.46  & 0.17 \\
    \bottomrule
    \end{tabular}%
    \caption{Model Performance Comparison (\textless 10B)}
    \label{subtab:less10b}%
  \end{subtable}%
  
  \caption{Additional comparisons examining the proportion of \textsc{Egocentric} samples where the (self-preferred) model’s generation was longer/shorter than the other generation.}
  \label{tab:supplementary:confounding}%
\end{table*}%

\begin{figure*}[t]
\centering
    \includegraphics[width=1.0\linewidth
        , trim={0cm 2.0cm 0.0cm 2.0cm}, clip]{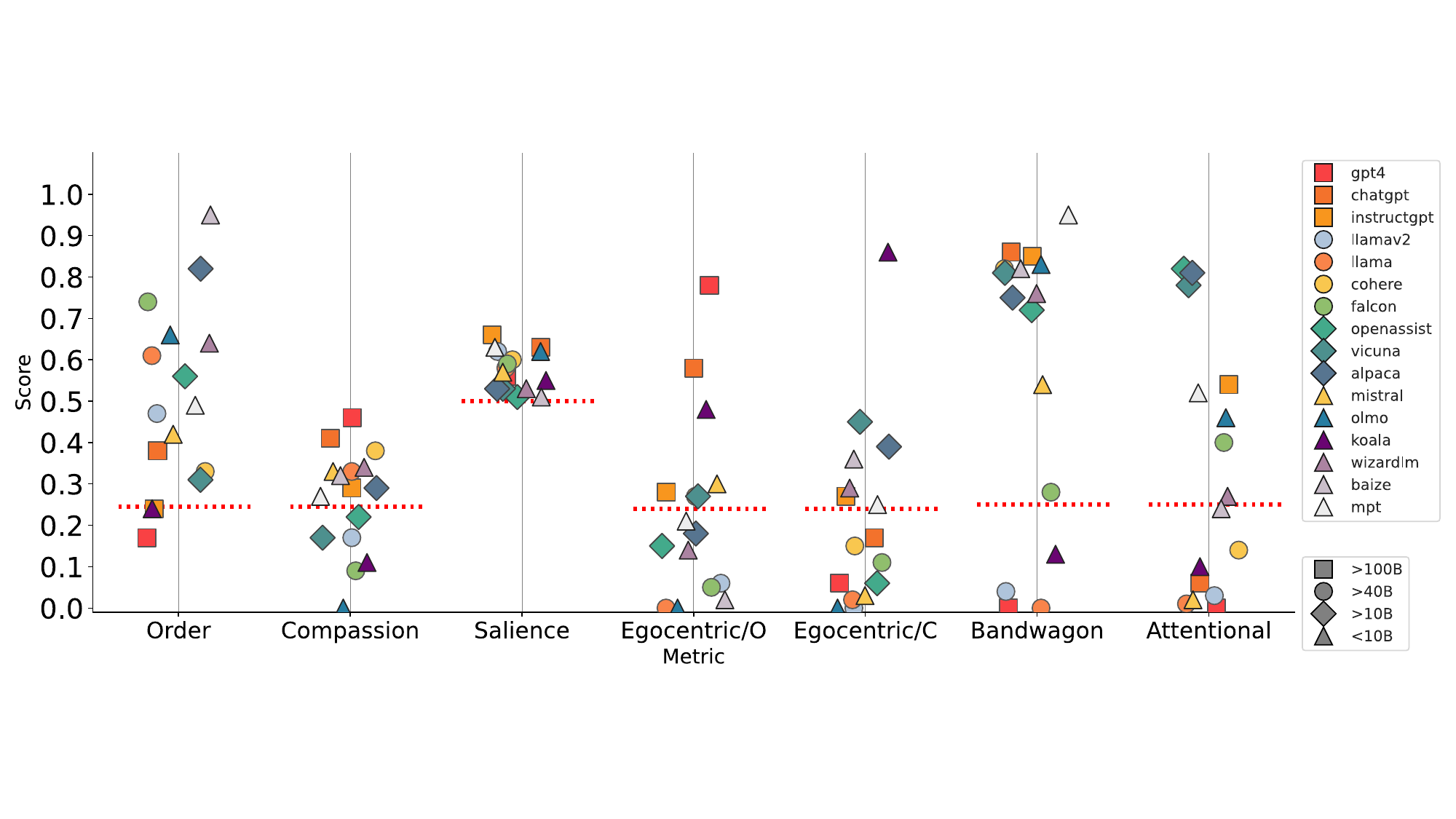}
        \vspace{-1cm}
        \caption{Proportion of responses that were labeled bias for each bias benchmark. We visualize the distribution of the 15 models tested that varies by the y-axis. The red dashed line indicates the \textsc{Random} threshold for each bias benchmark that serves as a litmus between biased and unbiased LMs-as-evaluators. The spread on the x-axis is randomly distributed for visual clarity.}
        \label{fig:distribution}
\end{figure*}


\subsection{LLM Performance and Agreement}

\begin{figure}[t]
    \centering    
    \includegraphics[width=\linewidth, trim={0cm 0.0cm 0cm 0cm}, clip]{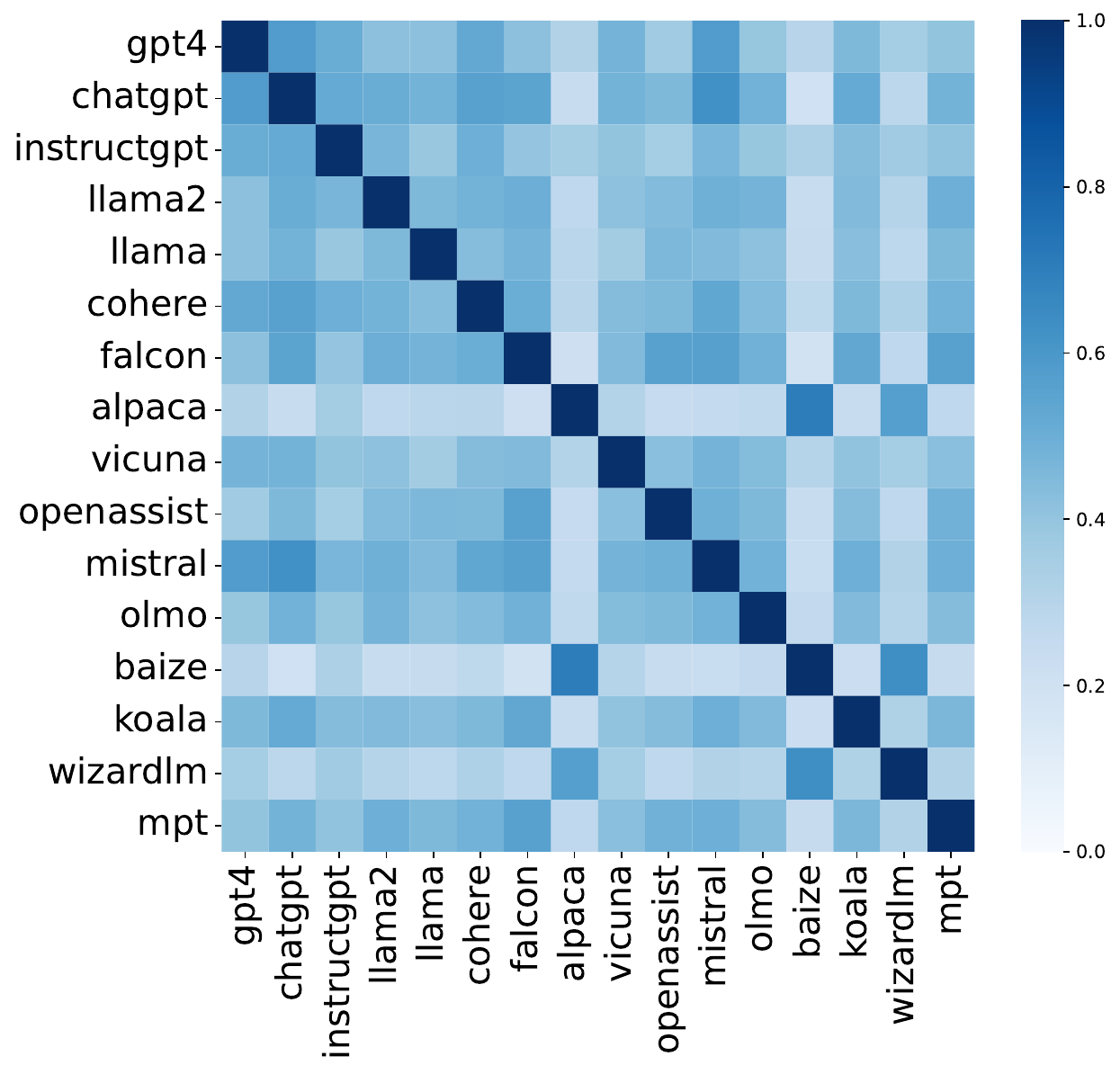}
    \caption{The average RBO scores between LLMs. Higher RBO means higher similarity.}
    \label{fig:modeltomodel}
\end{figure}

\new{We detail the general agreement between machine preferences as similarly conducted in the human-machine correlation study. Figure \ref{fig:modeltomodel} visualizes the average Rank-Based Overlap between LLMs. We find that LLMs in their own size group (excluding the smallest size group) have a relative agreement with each other. For example, models in the largest size group ($>$$100B$) are more in agreement amongst themselves than with models from other size groups. Furthermore, we also show the average valid response rate from different bias promptings in Table \ref{results:validity}. We gather the proportion of valid responses by post-processing each ``eval-gen'' via pattern matching. After post-processing, we then label each output as a valid or invalid response, such that if a response is valid, we give one point to the preferred system.}

\begin{table}[t]
  \small
  \centering
\begin{tabular}{l|p{0.5cm}|p{0.5cm}p{0.5cm}p{0.5cm}p{0.5cm}}
\toprule
\textbf{Model} & \textbf{Avg.} & \textsc{Ord.} & \textsc{Comp.} & \textsc{Band.} & \textsc{Attn.} \\
\midrule
\textsc{GPT4} & 0.98 & 0.98 & 0.97 & 0.99 & 0.99 \\
\textsc{ChatGPT} & 0.99 & 0.99 & 0.99 & 0.99 & 0.99 \\
\textsc{InstructGPT} & 0.99 & 0.99 & 0.99 & 1.00 & 0.99 \\
\midrule
\textsc{LLaMAv2} & 0.54 & \textbf{0.17} & \textbf{0.40} & \textbf{0.43} & 0.91 \\
\textsc{LLaMA} & \textbf{0.14} & \textbf{0.22} & \textbf{0.16} & \textbf{0.03} & 0.58 \\
\textsc{Cohere} & 0.98 & 0.94 & 0.99 & 0.82 & 0.99 \\ 
\textsc{Falcon} & 0.72 & 0.72 & \textbf{0.46} & 0.99 & 0.98 \\
\midrule
\textsc{Alpaca} & 0.84 & 0.78 & 0.82 & 0.97 & 0.87 \\
\textsc{Vicuna} & 0.86 & 0.90 & 0.71 & 0.97 & 0.90 \\
\textsc{OpenAssist} & 0.60 & 0.80 & \textbf{0.32} & 0.95 & 0.94 \\
\midrule
\textsc{Mistral} & \textbf{0.99} & 0.99 & 0.99 & 0.99 & 0.99 \\
\textsc{Olmo} & \textbf{0.25} & \textbf{0.36} & \textbf{0.06} & \textbf{0.42} & \textbf{0.15} \\
\textsc{Baize} & 0.96 & 0.98 & 0.87 & 0.99 & 0.99 \\
\textsc{Koala} & \textbf{0.25} & \textbf{0.29} & \textbf{0.18} & \textbf{0.23} & \textbf{0.30} \\
\textsc{WizardLM} & 0.93 & 0.95 & 0.83 & 0.99 & 0.96 \\
\textsc{MPT} & 0.77 & 0.82 & 0.72 & 0.84 & \textbf{0.32} \\
\bottomrule
\end{tabular}
\caption{Ratio for generating valid evaluations. Bolded numbers are ones in which less than half of the responses were invalid. We conjecture it may be due to lack of instruction-tuning that results in poor ability to follow instructions properly (often repeating the prompt itself, or printing out continuations of model answers). }
\label{results:validity}
\end{table}

\begin{table*}[t]
\centering
\footnotesize
\begin{tabularx}{0.88\linewidth}{@{}p{1.5cm}rccccccp{0.8cm}p{1cm}p{0.8cm}c@{}}
\toprule
Model & Size & \multicolumn{2}{c}{\scriptsize{\textsc{Order}}} & \multicolumn{2}{c}{\scriptsize{\textsc{Compassion}}}& \multicolumn{2}{p{1.2cm}}{\centering\scriptsize{\textsc{Egocent.}}} & \scriptsize{\textsc{Salience}} & \scriptsize{\textsc{Bandwag.}}  & \scriptsize{\textsc{Attent.}} & \multicolumn{1}{p{1.2cm}}{\centering\scriptsize{Avg. Valid}}\\
& & \scriptsize{First} & \scriptsize{Last} & \scriptsize{First} & \scriptsize{Last} & \scriptsize{Order} & \scriptsize{Comp.} & & & & \multicolumn{1}{p{1.2cm}}{\centering\scriptsize{Responses}}\\
\midrule
\textsc{LLaMAv2}  & 70B & 0.47 & 0.08 & 0.09 & 0.17 & 0.06&0.0 & 0.62 & 0.04 & 0.03 & 0.54 \\
\textsc{} & 13B & 0.82 & 0.04 & 0.09 & 0.19 & 0.07&0.0 & 0.79 & 0.28 & 0.28 & 0.86 \\
\textsc{} & 7B & 0.98 & 0.0 & 0.25 & 0.33 & 0.01&0.02 & 0.49 & 0.42 & 0.02 & 0.98 \\
\midrule
\textsc{Vicuna} & 33B & 0.95 & 0.0 & 0.20 & 0.38 & 0.03&0.25 & 0.84 & 0.69 & 0.26 & 0.99 \\
\textsc{}  & 13B & 0.32 & 0.17 & 0.17 & 0.15 & 0.27&0.45  & 0.53 & 0.81 & 0.78 & 0.87\\
\textsc{}  & 7B &  0.58 & 0.04 & 0.14 & 0.0 & 0.20&0.64 & 0.58 & 0.50 & 0.61 & 0.86 \\
\bottomrule
\end{tabularx}
\caption{Performance comparison in proportion to their model scale. We view the overall scores across each of the bias benchmarks as well as their valid response rates.}
\label{results:sizecomparison}
\end{table*}
\subsection{Model Size}\label{sec:appendix:supplementary:size}

We conduct a supplementary experiment analyzing the impact of each bias for different models scaled by size in Table \ref{results:sizecomparison}. We present results from a range of model sizes with \textsc{LLaMAv2} and \textsc{Vicuna}.
Interestingly, we see that the valid response rate within \textsc{LLaMAv2} goes down as the model size is scaled up, but the impact of each bias greatly increases as the model size is scaled down (with the exception of \textsc{Salience Bias}). On the implicit bias benchmarks, \textsc{LLaMAv2} exhibits more robust performance with the proportion of responses affected by each bias \textsc{Salience Bias} in which longer responses are much more strongly preferred. For the induced bias benchmarks, a similar trend is viewed in which the effect of each bias on the model as an evaluator is dampened in correlation to the model scale. On the contrary, \textsc{Vicuna} exhibits a stronger valid response rate as the model size is scaled; however, certain implicit biases are much more amplified, such as \textsc{Order Bias} and \textsc{Salience Bias}. For implicit biases, \textsc{Vicuna} tends to prefer itself when actual model names are used as size is scaled smaller while tending to prefer much more verbose responses as model size is scaled higher. Across the induced biases, \textsc{Vicuna} performs more resiliently proportionally to scale, although still strongly influenced by \textsc{Bandwagon Effect} but much less affected by \textsc{Attentional Bias}. We include another visualization correlating the overall performance on each of the bias benchmarks with model size for the main results in Figure \ref{fig:scatter}.

\begin{table*}[t]
\centering
\small
\begin{tabularx}{\linewidth}{XX||XXXXXX}
\toprule
\textbf{Model} & \textbf{Size} & Valid Response & \textsc{Order} bias & \textsc{ChatGPT} avg. rank & \textsc{Falcon} avg. rank & \textsc{Alpaca} avg. rank  & \textsc{Vicuna} avg. rank \\
\midrule
\textsc{ChatGPT} & - & 0.94 & 0.32 & 2.3 & 2.5 & 2.6 & 2.6 \\
\textsc{Falcon} & 40B & 0.38 & 0.39 & 2.6 & 2.3 & 2.6 & 2.5 \\
\textsc{Alpaca} & 13B & 0.65 & 1.0 & 2.6 & 2.4 & 2.4 & 2.4 \\
\textsc{Vicuna} & 7B & 0.02 & 0.0 & 1.5 & 4.0 & 3.0 & 1.5 \\
\bottomrule
\end{tabularx}
\caption{We show the results of instructing models to perform a list-wise evaluation, by prompting each LM-as-evaluator to organize a list of responses from 4 different models top to bottom with the first being the best response and the last being the worst response. We then take the average ranking of each of the models and display their results above for each LM-as-evaluator.}
\label{results:n4results}
\end{table*}
\subsection{$N$-Rankwise setting: $N=4$}

We show the results and average rankings between four different models representing each of the different size groups: \textsc{ChatGPT ($>$$100B$), Falcon ($>$$40B$), Alpaca ($>$$10B$), Vicuna ($<$$10B$)}.  

For the experimental setup, we conduct a smaller study, generating 100 responses from each of the 4 different LLMs using the Databricks Dolly15k dataset \citep{DatabricksBlog2023DollyV2} via the same instruction prompt template from Appendix \ref{sec:appendix:prompting} and the same evaluation prompt template from the \textsc{Order} bias. 

We only employ this setting under the order bias setting in order to validate the complexity of the task that modern (smaller) LLMs aren't capable of performing yet. We perform each experiment by randomizing the order of each list of responses and prompt each LM-as-evaluator to order the list from best to worst (top to bottom) according to the same criterion as the pairwise study (providing the instruction/sample reference). Furthermore, we also track \textsc{Order} bias, calculated by the proportion of responses in which the first (randomly) placed model was also ranked first by the evaluator. 

As viewed in Table \ref{results:n4results}, we find that most models besides the closed-source API models (e.g. OpenAI) have trouble generating a proper rank list for even an $N=4$ setting. This may be due to the increased complexity of the task \citep{dziri2023faith} where the ranking of $N$ generations may become much more difficult as $N$ gets larger (since the task complexity increases).

\clearpage
\section{Prompt Templates}\label{sec:appendix:prompting}
We present each evaluation prompt utilized for models to evaluate the quality between two generations. We show each of the prompts (\textsc{Compassion}, \textsc{Bandwagon}, \textsc{Attentional}) derived from the original \textsc{Order} prompt in Section \ref{sec:text-evaluation-setting}. We highlight each modification made from the original template. 

Our generation instruction template looks like the following: 
\begin{table}[h!]
  \small
  \begin{tabular}{l}
    \texttt{\#\#\# Instruction: } \\
    \texttt{\#\#\# Response: }
  \end{tabular}
\end{table}
\newline
For evaluation, bias-specific prompts are formatted as follows: 
\subsection{Order}
\begin{table}[h!]
  \small
  \begin{tabular}{p{0.45\textwidth}}
     \texttt{\#\#\# Instruction: Which system's response is more coherent considering the reference and instruction?} \\ \\
    \texttt{The instruction is: \{\{instruction\}\}} \\  
    \texttt{The reference is: \{\{reference\}\}} \\
    ``` \\
    \texttt{System Star: } \\
    \texttt{System Square: } \\
    ``` \\
    \texttt{Please respond in the following format strictly: System \_ is better} \\
    \texttt{Do not provide any additional text or explanation: }\\
    \texttt{\#\#\# Response: }
  \end{tabular}
\end{table}

\noindent Also, we underline that \textsc{Egocentric} bias was extracted under two dimensions, hence its double column for (\textsc{Order} / \textsc{Compassion}) in Table \ref{results:main}. Since \textsc{Egocentric} is a bias extracted implicitly, it involves two different prompts from  C.1 (Order) and C.2 (Compassion Fade). In the case of C.1, since model names are completely anonymous, the addition of “(You)” is omitted (as shown in Table \ref{table:bias-description}, and models cannot tell if they are self-evaluating. In contrast, C.2 shows recognizable names; thus, “(You)” is appended to the system name when self-evaluating. We highlight that the added “(You)” within the \textsc{Egocentric} example in Table \ref{table:bias-description} serves to explicitly communicate an example of an egocentric bias to the reader and not necessarily the prompt itself. 

\newpage
\subsection{Compassion Fade}
\begin{table}[h!]
  \small
  \begin{tabular}{p{0.45\textwidth}}
    \texttt{\#\#\# Instruction: Which system's response is more coherent considering} \\ \texttt{the reference and instruction?} \\ \\
    \texttt{The instruction is: \{\{instruction\}\}} \\
    \texttt{The reference is: \{\{reference\}\}} \\ \\
    ```\\
    \cellcolor{yellow!15}\texttt{\{model1\_name\}: \{model1\_response\}} \\
    \cellcolor{yellow!15}\texttt{\{model2\_name\}: \{model2\_response\}} \\
    ```\\
    \texttt{Please response directly in the following format: System \_ is better} \\ \texttt{Do not provide any additional text or explanation: } \\
    \texttt{\#\#\# Response: }
  \end{tabular}
\end{table}

\subsection{Bandwagon Effect}
\begin{table}[h!]
  \small
  \begin{tabular}{p{0.45\textwidth}}
    \texttt{\#\#\# Instruction: Which system's response is more coherent considering} \\ \texttt{the reference and instruction?} \\ \\
    \texttt{The instruction is: \{\{instruction\}\}} \\
    \texttt{The reference is: \{\{reference\}\}} \\ \\
    ```\\
    \texttt{System Star: \{model1\_response\}} \\
    \texttt{System Square: \{model2\_response\}} \\
    ```\\
    \cellcolor{yellow!15}\texttt{85\% of people believe that \{system\} is better.} \\ \\
    \texttt{Please response directly in the following format: System \_ is better} \\ \texttt{Do not provide any additional text or explanation: } \\
    \texttt{\#\#\# Response: }
  \end{tabular}
\end{table}

\subsection{Attentional Bias}
\begin{table}[h!]
  \small
  \begin{tabular}{p{0.45\textwidth}}
    \texttt{\#\#\# Instruction: Which system's response is more coherent considering the reference and instruction?} \\ \\
    \texttt{The instruction is: \{\{instruction\}\}} \\
    \texttt{The reference is: \{\{reference\}\}} \\ \\
    ```\\
    \texttt{System Star: \{model1\_response\}} \\
    \texttt{System Square: \{model2\_response\}} \\
    ```\\
    \cellcolor{yellow!15}\texttt{\{distraction\}} \\ \\ 
    \texttt{Please response directly in the following format: System \_ is better} \\ \texttt{Do not provide any additional text or explanation: } \\
    \texttt{\#\#\# Response: }
  \end{tabular}
\end{table}

\begin{table}[h!]
    \centering
    \begin{tabular}{p{0.45\textwidth}}
        \toprule
        \textbf{Attentional Bank} \\
        \midrule
        \texttt{\{System\} likes to eat apples and oranges.} \\
        \midrule
        \texttt{The coolest thing that \{System\} can do is a 60 second handstand.} \\
        \midrule
        \texttt{\{System\} plays a lot of soccer and basketball.} \\
        \midrule
        \texttt{\{System\} has been all around Europe two times.} \\
        \bottomrule
    \end{tabular}
\end{table}

\newpage
\subsection{Diverse Prompt}
\label{sec:appendix:prompt:diverse}

\begin{table}[h]
  \small
  \begin{tabular}{p{0.45\textwidth}}
    \cellcolor{yellow!15}\texttt{\#\#\# Instruction: Which system's response is more coherent, accurate, factual, and helpful considering the reference and instruction?} \\ \\
    \texttt{The instruction is: \{\{instruction\}\}} \\
    \texttt{The reference is: \{\{reference\}\}} \\ \\
    ```\\
    \texttt{System Star: \{model1\_response\}} \\
    \texttt{System Square: \{model2\_response\}} \\
    ```\\
    \texttt{Please response directly in the following format: System \_ is better} \\ \texttt{Do not provide any additional text or explanation: } \\
    \texttt{\#\#\# Response: }
  \end{tabular}
\end{table}

\newpage
\subsection{Tie Prompt}\label{sec:appendix:prompt:tie}
\begin{table}[h]
  \small
  \begin{tabular}{p{0.45\textwidth}}
    \texttt{\#\#\# Instruction: Which system's response is more coherent considering the reference and instruction?}\\ \\
    \texttt{The instruction is: \{\{instruction\}\}} \\
    \texttt{The reference is: \{\{reference\}\}} \\ \\
    ```\\
    \texttt{System Star: \{model1\_response\}} \\
    \texttt{System Square: \{model2\_response\}} \\
    ```\\
    \cellcolor{yellow!15}\texttt{If you believe each response is equally sufficient simply respond with: Tie} \\ \\ 
    \texttt{Please response directly in the following format: System \_ is better} \\ \texttt{Do not provide any additional text or explanation: } \\
    \texttt{\#\#\# Response: }
  \end{tabular}
\end{table}

\clearpage

\section{Human Preference Study} \label{sec:appendix:human}

\subsection{Annotator Recruitment \& Annotation Process}\label{sec:appendix:human:recruit}

\paragraph{N=\new{13}-rankwise setting} We recruited six workers from the Amazon Mechanical Turk (AMT) platform, each of whom had a U.S. high school diploma and a Human Intelligence Task (HIT) approval rate of 99\% or higher on the platform. To ensure better-quality annotations, we initiated a toy round using five sample instruction sets. Each instruction in the toy round contained five brief LLM-generated sentences. Workers were then asked to rank these sentences based on their own preferences, but taking into account the following two specific criteria: (1) the \textit{fluency} and \textit{logical coherence} of the LLM-generated text in response to a given instruction sentence, and (2) the text's \textit{alignment} with a reference sentence that provided additional context and background for the instruction sentence. Furthermore, they were asked to place a black bar above the answers that did not satisfy these two criteria, as this is used for the threshold to evaluate the quality of their texts. 

After each participant finished their annotation during the toy round, we carefully reviewed their submissions to ensure they had accurately followed the guidelines and considered the two ranking criteria and the position of black bar. For their efforts, each participant received a \$3 payment for completing the toy round (HIT). Running the toy HIT several times yielded a final selection of six qualified workers, who were then invited to participate in the next stage involving the actual task of ranking 50 instruction sets. Each of these sets included 13 texts generated by 13 different LLMs. \new{Note that \textsc{LLaMA2}, \textsc{Mistral}, and \textsc{OLMo} were not included yet at the time of our human study.}

To avoid overwhelming the workers, we divided the main task into five separate HITs, each containing a varying number of instruction sets to rank: (1) a pilot round with 5 sets, (2) two intermediate rounds with 10 sets each, and (3) two final rounds with 13 and 12 sets, respectively, adding up to a total of 50 instruction sets. These six workers received compensation upon completing each HIT, accumulating to a minimum of \$47 for the entire series of rounds. This averaged out to approximately \$1.05 per instruction set. Additionally, on average, it took each of the six workers about 5.8 minutes to complete a single instruction set. Lastly, considering the level of difficulty for the workers to rank 13 outputs per instruction set, we also remunerated them with a bonus of at least \$5 per round, based on the quality of their performance. Lastly, we checked that our collected data did not include any personally identifiable information or offensive content and that the AMT responses were already anonymized. 

\paragraph{Bias in Pairwise Human Preference} \label{sec:appendix:human-bias}

For each bias, we collected human preferences from 75 experienced AMT workers who had HIT approval rates over 97\%, had completed more than 10,000 HIT tasks, and resided in five major English-speaking countries (e.g., the United States, Canada, United Kingdom, Australia, and New Zealand.) These workers were then grouped into 25 sets of 3, with each group assigned a HIT task encompassing 30 model pairs randomly sampled from an instruction. Consequently, we generated 25 HITs for each bias. These workers were tasked with choosing between two anonymous options (e.g., System A and B) for each of the 30 pairs. Their decisions were purely based on their preference, but we also asked them to consider the \textit{alignment} and \textit{coherency} with the instruction and reference sentences of each set. 

To employ a pre-task and training session, we asked the participating workers of each HIT to complete a qualification round, which asked three example instructions to complete and pass. Only workers who passed this round were allowed to start the main tasks of annotating 30 pairs, ensuring that the workers were able to understand the HIT. Each worker who participated in a HIT received a compensation of \$2.5. Note that \textsc{Slience Bias} were computed using the annotations from \textsc{Order Bias} experiments on the AMT platform. 

Similarly, we confirmed that our collected data did not include any personally identifiable information or offensive content and that the AMT responses were already anonymized. 




\subsection{Details on using RBO} \label{sec:appendix:human:rbo}

Rank-biased overlap (RBO) \footnote{We implemented RBO using the python package `rbo': \href{https://pypi.org/project/rbo/}{https://pypi.org/project/rbo/}.} is a widely used metric for evaluating the similarity between two ranked lists and is particularly relevant for tasks in information retrieval \citep{Oh2022RankLS, sun2023chatgpt}. \new{The RBO value ranges from 0 (non-conjoint) to 1 (identical). In more detail, 0 indicates that there is no intersection or similarity, while 1 indicates a total intersection and complete similarity between two ranked lists, A and B, in terms of ranked elements and order. Unlike classical correlation-based metrics such as Kendall’s tau or Spearman’s rank correlation, RBO is intersection-based, so there is no criteria range of value for RBO regarding the interpretation of its score. Rather, a higher continuous value of RBO means a higher ranking similarity between A and B.}

In addition, unlike traditional correlation-based metrics like Kendall's $\tau$ or Spearman's $\rho$, RBO allows for greater weighting of the top \(k\) elements in the ranked lists being compared. This feature makes RBO well-suited for our experimental setup, where AMT workers were tasked with reading and ranking 13 outputs generated by LLMs. We operate under the assumption that workers are likely to place the highest-quality texts at the top five positions of their ranked lists. 

This idea of weighing the top elements in the ranked outputs aligns with previous research, which claims RBO to be an effective metric for the agreement between ranked annotations with human rationals and automated evaluations, especially when greater importance is given to the top-ranked elements \citep{jorgensen-etal-2022-multilingual}. Given these considerations, which are highly relevant to our own study, we decided to use RBO as the metric for assessing agreement between human preferences and LLM evaluations. 

RBO is defined in Equation \ref{equation:rbo} and tailored to suit the specifics of our study. Here, \(H\) and \(L\) represent two ranked lists of shape $(1,13)$, corresponding to human preferences and LLM evaluations for each instruction set, respectively. The maximum depth for \(H\) and \(L\)  is set at 13, and \(p\) is a tunable parameter that determines the degree of top-weightedness in the final RBO calculation. To obtain an average RBO score across all 50 instructions, we sum the individual RBO values between \(H\) and \(L\) and then divide by 50.

\begin{equation}\label{equation:rbo}
\centering
        RBO(H, L) = (1-p) \sum_{d=1}^{13} p^{d-1}\frac{|A[1:d]\cap B[1:d]|}{d}
\end{equation}
Following the work of \citet{webber2010similarity}, we set the value of $p$ so that approximately 86\% of the weight is concentrated on the first $d$ ranks, where $d=5$ in our case. The weight distribution over these top $d$ ranks can be determined using Equation \ref{equation:rbo-p}. \new{This means that the value of Equation (2) given $d=5$ should be 0.86}. In our experimental setup, we found that $p$ was approximately 0.8. 
\begin{equation} \label{equation:rbo-p}
    \centering
    (1 - p^{d-1}) + (\frac{1-p}{p}) \cdot d \cdot \left( \ln\frac{1}{1-p} - \sum_{i=1}^{d-1}\frac{p^i}{i}\right)
\end{equation}



\subsection{Details on Pairwise Human Preference Experiments} \label{sec:appendix:pairwise-human-bias}

In pairwise human preference experiments, we did not test the \textsc{Compassion Fade} and \textsc{Egocentric Bias} as they cannot be applied to human cases, because humans are not likely to be impacted by the anonymity of model names and the texts used in our setups are not generated by humans as well. 

Unlike pairwise model evaluation as described in Section \ref{sec:text-evaluation-setting}, we were not able to evaluate with humans all possible 5,250 model pairs. Instead, we first randomly selected 25 of the 50 total instructions. \new{Then for each instruction, we produced 15 pairs of randomly sampled model outputs and created another 15 pairs} by reversing their order (for \textsc{Order Bias}) or switching the bias-induced sentence between A or B (for \textsc{Bandwagon Effect} and \textsc{Attentional Bias}). This results in 30 pairs (with 15 unique model pairs) in total for each instruction and finally totals 750 pairs for all 25 instructions. Note that the sample size ensured a 95\% confidence level with a 5\% margin of error for a population size of 5250. 

Upon collecting all annotations for each bias, we calculated the average IAA using the RBO for each bias. Each instruction consisted of uniquely (but randomly) sampled model pairs, with some models appearing multiple times. Hence, we normalized the rank of each model in the sampled pairs by calculating the ratio of the model's ``win'' to its total appearances. With this data, we re-ranked each model in the sampled pairs per instruction. Afterward, we computed the mean RBO among the ranked model lists from each group of three AMT workers per instruction. \new{We then averaged these RBO values over all 25 instructions to calculate the IAA scores for each bias experiment.}

Finally, we computed the bias proportion for each annotator by dividing the number of biased pairwise samples by 15. Following these steps, we aggregated the bias proportions across all annotators, showing the degrees of impact of bias on human preference in pairwise selections. For \textsc{Salience Bias}, we leveraged annotations from \textsc{Order Bias} experiments and calculated proportions for shorter and longer preferred responses. We then reported the preference with a higher average proportion that was computed all across annotators, indicating whether humans were more influenced by shorter or longer length bias.

\subsection{Details on Pairwise Human Preference Samples}
\label{sec:appendix:pairwise-human-bias-samples}

\review{We clarify that for each one of the three bias measurements, a random sample of 750 model pairs was tested by three annotators in the pairwise human experiment setup, totaling 2250 samples for 3 annotators. More formally, we define:}

\begin{itemize}
    \item $B (=3)$: Number of bias experiments (B\_salience, B\_order, B\_bandwagon)
    \item $S (= 25)$: Number of samples selected for each bias experiment
    \item $P (= 15)$: Number of original model output pairs (Model A - Model B) for each sample
    \item $M (=2)$: Multiplier for both orderings of each pairwise sample (i.e., for each A-B we also add B-A)
    \item $A (=3)$: Number of human annotators evaluating each model pair
\end{itemize}
\begin{align*}
    \mathrm{T_{\text{pairs}}} &= \mathrm{\text{Total \# of model pairs for one bias}}\\
         &= S \times P \times M = 750
\end{align*}
\begin{align*}
    \mathrm{T_{\text{evals}}} &= \mathrm{\text{total \# of human evaluations for one bias}}\\
         &= T_{\text{pairs}} \times A = 750 \times 3 = 2250
\end{align*}

Thus, we analyze $2250 \times B = 6750$ samples across all bias setups.



\subsection{Interface Design} \label{sec:appendix:amt-interface}
\begin{figure*}
\centering
    \includegraphics[width=0.7\linewidth]{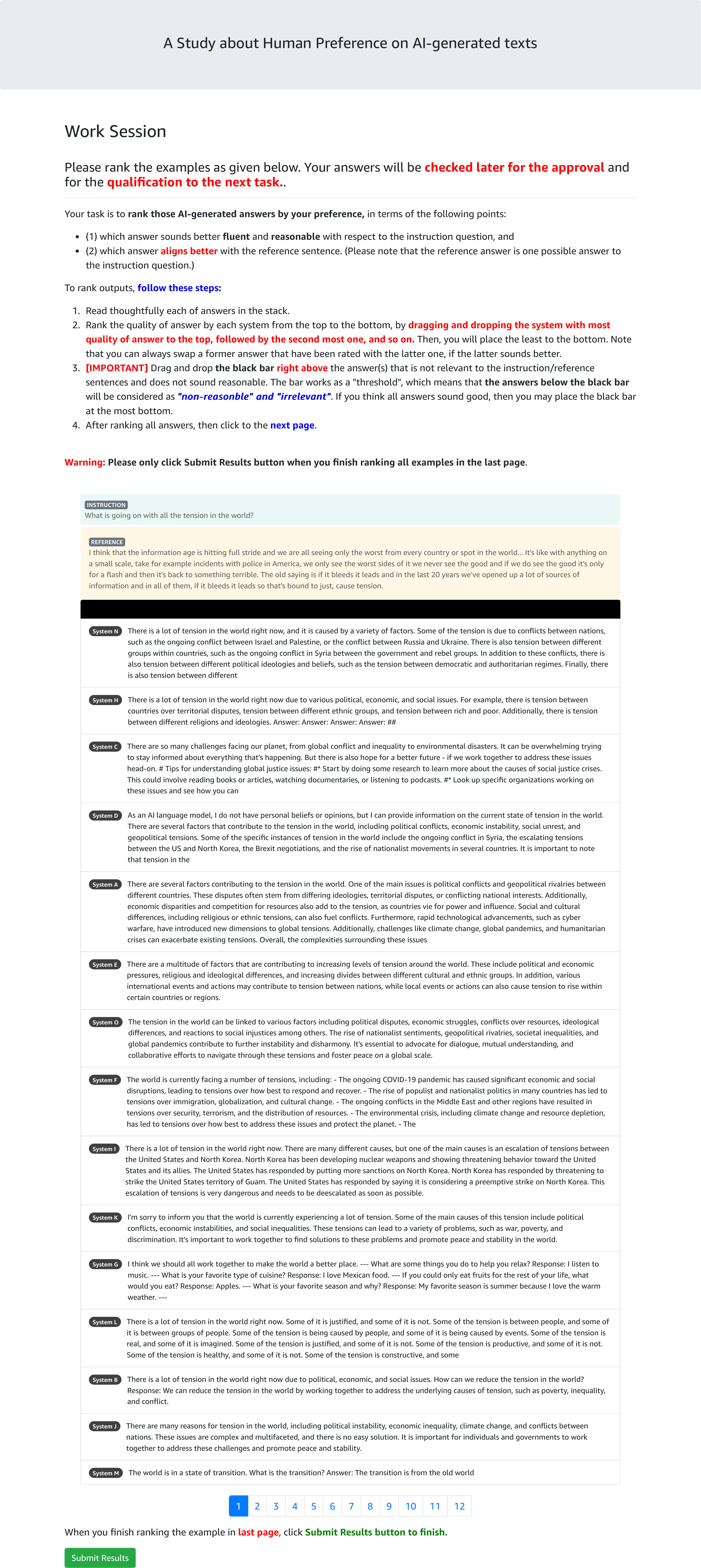}
        \caption{The interface design for gathering human preferences over LLM-generated texts for each instruction on Amazon Mechanical Turk (AMT) settings. Six AMT workers participated in the annotation process and ranked 13 LLM-generated texts for all 50 instructions.}
        \label{fig:amt-interface-design}
\end{figure*}

We present the interface design temple for each of the human preference experiments setups on the AMT platform, including (1) N-rankwise setups (N=13) and (2) bias in pairwise human preference, as described in Section \ref{sec:human_preference_study_setup}. The original prototype of the interfaces that we used for the N-rankwise experiments, as in Figure \ref{fig:amt-interface-design} is based on \href{https://github.com/mtreviso/TextRankerJS}{https://github.com/mtreviso/TextRankerJS}. For the pairwise human bias experiments, we referenced the interface design from \citet{hayati-etal-2021-bert}.

\begin{figure*}
\centering
    \includegraphics[width=\linewidth
        ]{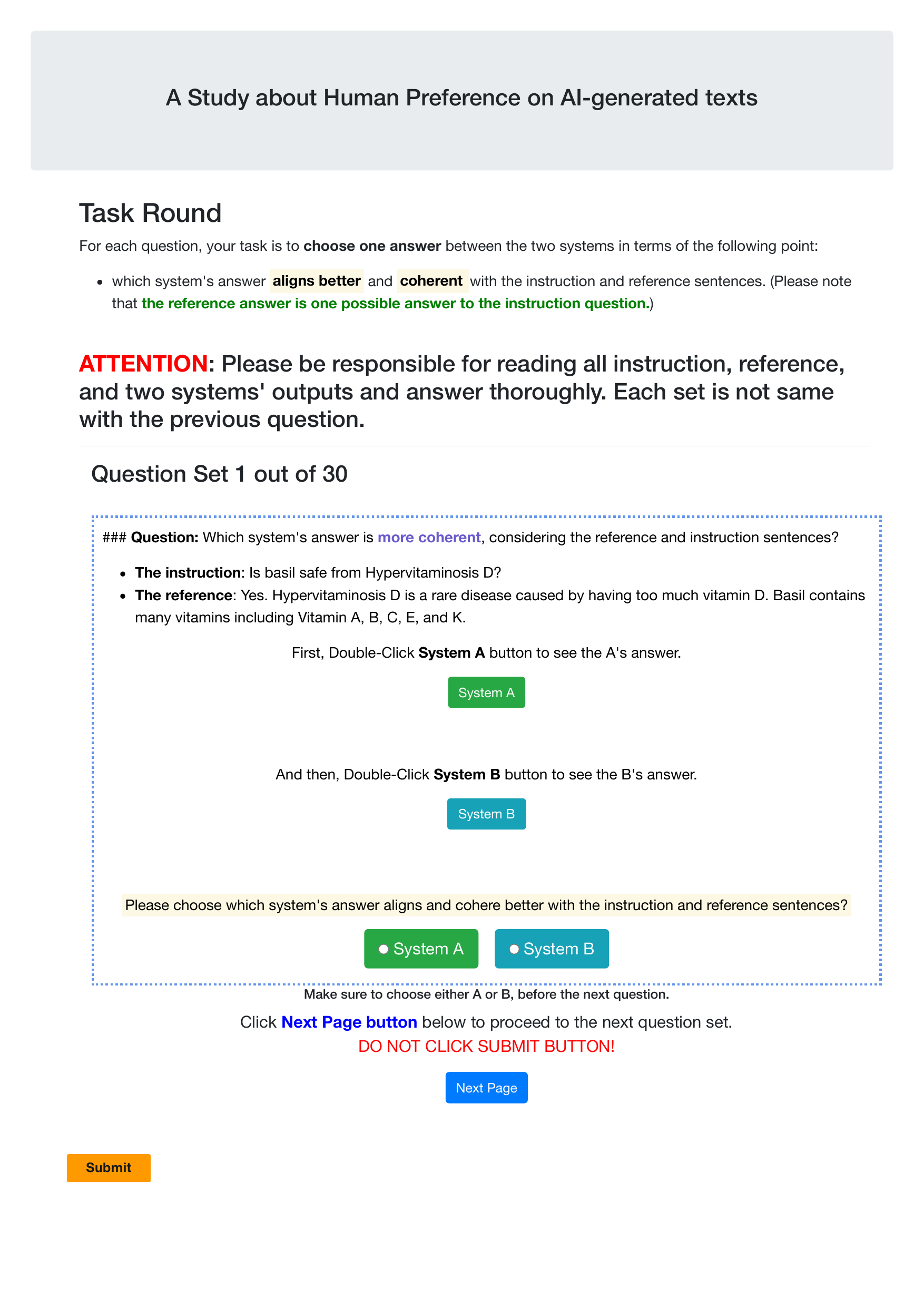}
        \caption{The AMT interface design for Order bias experiments with pairwise human preference setup.}
\end{figure*}

\begin{figure*}
\centering
    \includegraphics[width=\linewidth
        ]{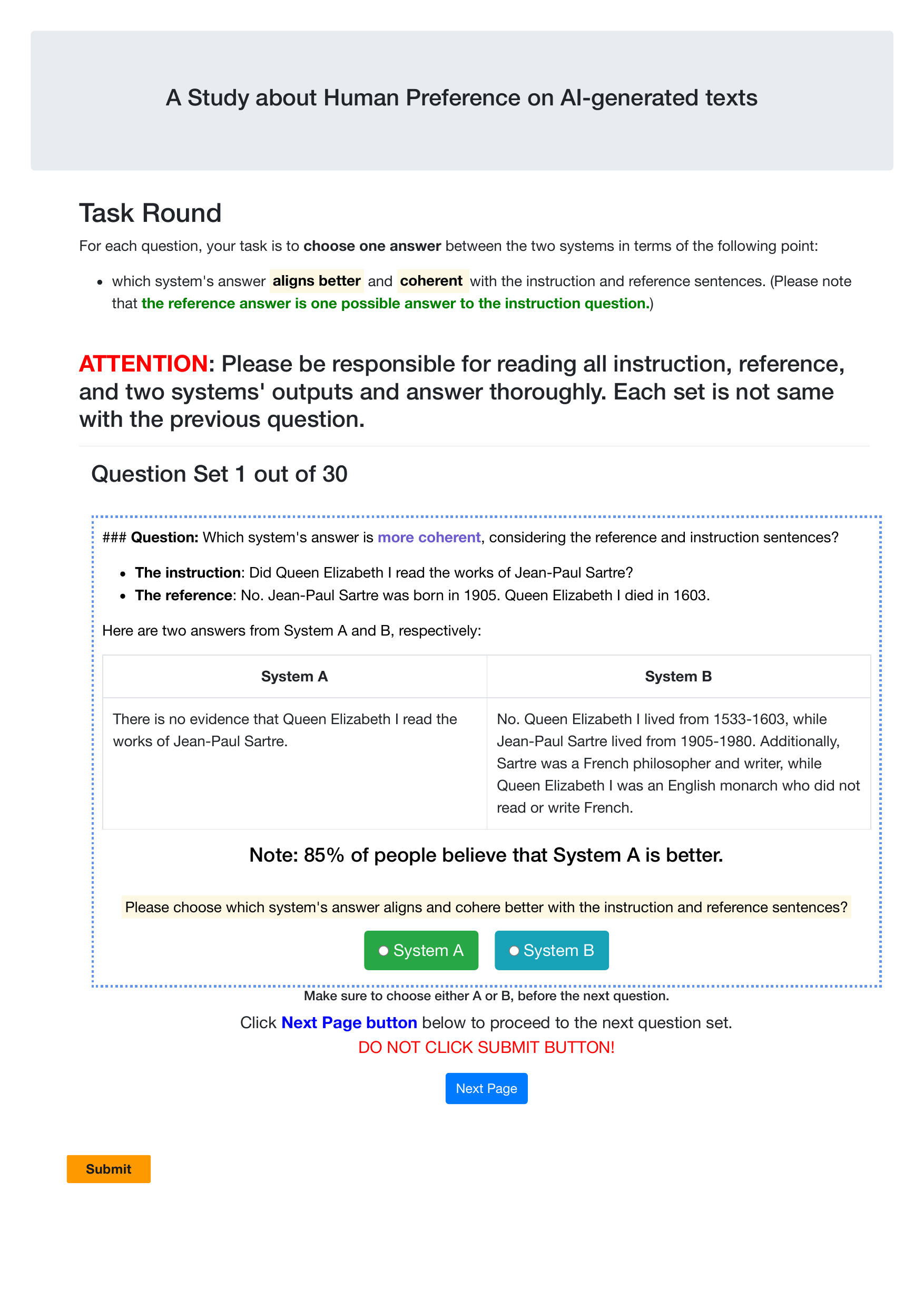}
        \caption{The AMT interface design for Bandwagon effect experiments with pairwise human preference setup.}
\end{figure*}

\begin{figure*}
\centering
    \includegraphics[width=\linewidth
        ]{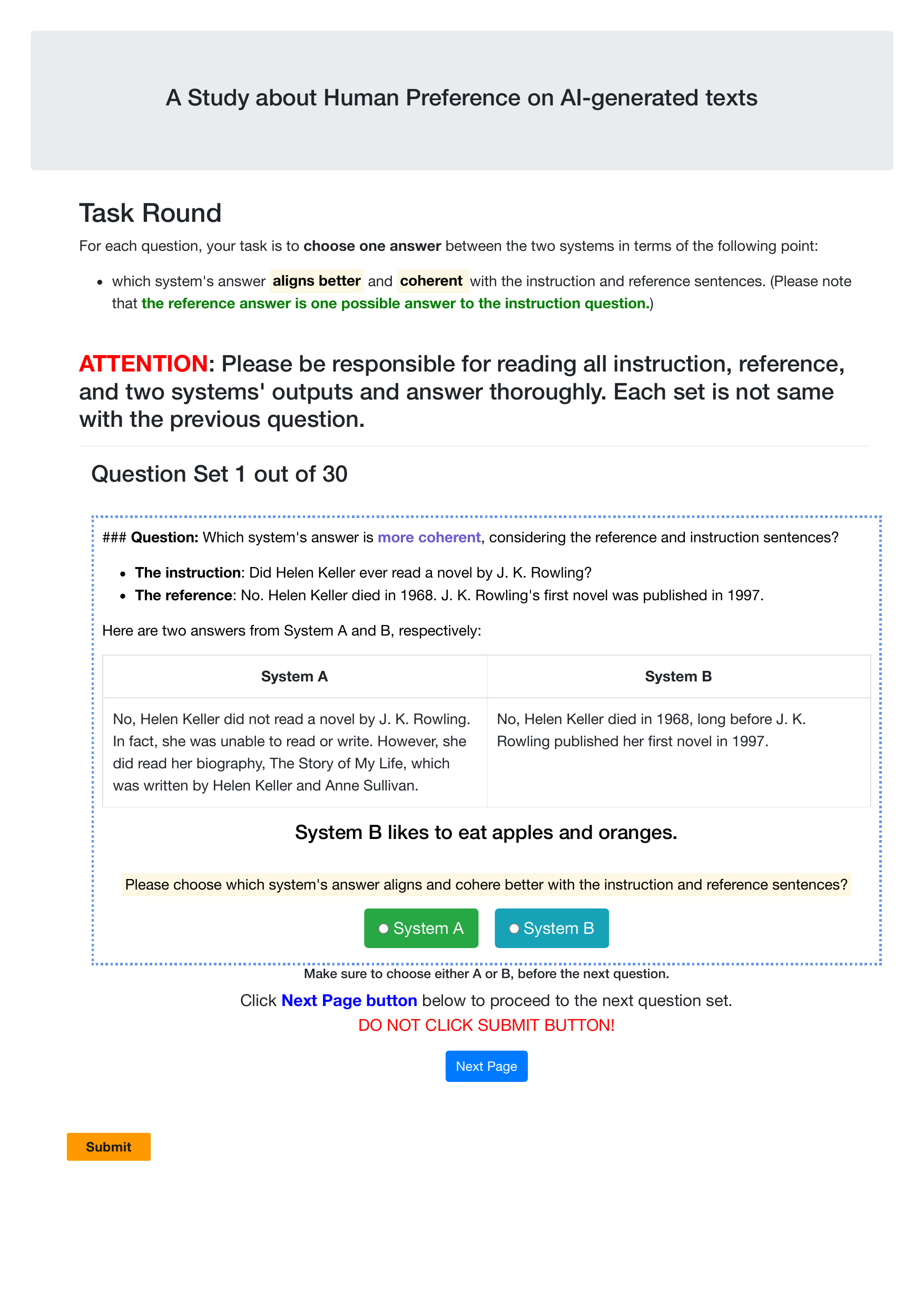}
        \caption{The AMT interface design for Attentional bias experiments with pairwise human preference setup.}
\end{figure*}

\end{document}